\documentclass[lettersize,journal]{IEEEtran}
\usepackage{amssymb,amsmath,amsfonts,bm}
\usepackage{algorithmic}
\usepackage{algorithm2e}
\usepackage{array,makecell}
\usepackage{graphicx}
\usepackage{epstopdf}
\usepackage{epsfig}
\usepackage[caption=false,font=normalsize,labelfont=sf,textfont=sf]{subfig}
\usepackage{multirow,multicol}
\usepackage{float}
\usepackage{diagbox}

\usepackage{cite}
\usepackage{adjustbox}
\usepackage{cleveref}
\crefname{Assumption}{assumption}{assumptions}

\graphicspath{{figures/}}
\allowdisplaybreaks

\usepackage[dvipsnames]{xcolor}
\usepackage{xspace}
\newcommand{\comment}[1]{}

\usepackage{comment}

\newcommand{\av}{\mathbf{a}}

\newcommand{\xv}{\mathbf{x}}

\newcommand{\yv}{\mathbf{y}}
\newcommand{\zv}{\mathbf{z}}

\newcommand{\Yv}{\mathbf{Y}}
\newcommand{\Zv}{\mathbf{Z}}

\newcommand{\argmin}{\mathop{\mbox{\rm arg\,min}}}

\newcommand{\vect}[1]{\mathbf{#1}}

\newcommand{\norm}[1]{\left \| #1 \right \|}

\newcommand{\Expect}{\mathbb{E}}

\newcommand{\pth}[1]{\left( #1 \right)}
\newcommand{\qth}[1]{\left[ #1 \right]}
\newcommand{\sth}[1]{\left\{ #1 \right\}}

\newcommand{\iprod}[2]{\left \langle #1, #2 \right\rangle}

\newtheorem{Proposition}{Proposition}

\newtheorem{Assumption}{Assumption}

\ifodd 0
\newcommand{\tmcr}[1]{{\color{red}#1}}
\else
\newcommand{\tmcr}[1]{#1}
\fi

\newcommand{\mypara}[1]{{\smallskip \noindent \bf #1}\hspace{0.1in}}

\def \alg{\texttt{CSE-FSL}\xspace}

\begin{document}

\title{Federated Split Learning with Improved \\ Communication and Storage Efficiency}

\author{Yujia Mu,  Cong Shen,~\IEEEmembership{Senior Member,~IEEE}
\thanks{
	The work was supported in part by the U.S. National Science Foundation (NSF) under awards ECCS-2332060, ECCS-2143559, CPS-2313110, and CNS-2002902, and in part by the Commonwealth
Cyber Initiative (CCI) of Virginia under Award VV-1Q23-005.  A preliminary version of this work was presented at the 2023 IEEE International Conference on Communications \cite{mu2023icc}.
	}
\thanks{
            The authors are with the Charles L. Brown Department of Electrical and Computer Engineering, University of Virginia, Charlottesville, VA 22904, USA. Email: \texttt{\{ym8ct,cong\}@virginia.edu}.
            }
}



\maketitle

\begin{abstract}
Federated learning (FL) is one of the popular distributed machine learning (ML) solutions but incurs significant communication and computation costs at edge devices. Federated split learning (FSL) can train sub-models in parallel and reduce the computational burden of edge devices by splitting the model architecture. However, it still requires a high communication overhead due to transmitting the smashed data and gradients between clients and the server in every global round. Furthermore, the server must maintain separate partial models for every client, leading to a significant storage requirement. To address these challenges, this paper proposes a novel communication and storage efficient federated split learning method, termed \alg, which utilizes an auxiliary network to locally update the weights of the clients while keeping a \emph{single} model at the server, hence avoiding frequent transmissions of gradients from the server and greatly reducing the storage requirement of the server. Additionally, a new model update method of transmitting the smashed data in selected epochs can reduce the amount of smashed data sent from the clients. We provide a theoretical analysis of \alg, rigorously guaranteeing its convergence under non-convex loss functions. The extensive experimental results further indicate that \alg achieves a significant communication reduction over existing FSL solutions using real-world FL tasks.
\end{abstract}


\section{Introduction}
\label{sec:intro}    
\IEEEPARstart{A}{s} an emerging distributed machine learning (ML) paradigm, federated learning (FL) \cite{mcmahan2017communication} enables distributed clients to collaboratively train ML models without uploading their sensitive data to a central server. While FL addresses privacy concerns by keeping data localized, most existing FL approaches assume that clients possess sufficient \emph{computational} and \emph{storage} resources to perform local updates on large (potentially deep) models. However, this assumption breaks down in scenarios where clients, such as mobile and Internet-of-Things (IoT) devices, are resource-constrained. Consequently, these clients struggle to handle the heavy computational and storage demands of training deep ML models, making FL impractical in such settings.

To address this issue, split learning (SL) \cite{gupta2018distributed,vepakomma2018split,vepakomma2019reducing} is proposed. The basic SL approach splits the ML model into two parts. The first few layers are trained at the client, while the remaining layers are only stored at the server. Since a client only needs to train the first few layers of the model, the computation and storage requirements for the client is reduced compared with FL, while still allowing for the collaborative training of a deep ML model. 
Although SL reduces client-side resource requirements, it suffers from increased time delays and substantial communication overhead due to frequent interactions between the client and server for forward and backward pass computations.

In response to these limitations, federated split learning (FSL) \cite{thapa2020splitfed, abedi2023fedsl, he2020group} emerges as a hybrid solution, combining the parallelism of FL without sharing data with the model-splitting benefits of SL. FSL reduces the computational demands on clients more than FL while alleviating some delays seen in SL. However, despite these improvements, the underlying model partitioning still incurs significant communication costs, as both forward signals (smashed data) and backward gradients must be exchanged \textit{in each learning round}. A solution to reduce the communication cost is the local loss-based training method proposed in \cite{han2021accelerating}, which updates the client-side model locally without waiting to receive the gradients from the server, thereby reducing communication frequency. This architecture is most suitable for scenarios in which the server has sufficient computing power and storage, as it requires maintaining multiple copies of the model for each client. However, the resource consumption of the server is \emph{linearly} proportional to the number of clients in \cite{han2021accelerating}, which leads to scalability issues. It is thus challenging for the server to support the ML model training when many users participate in FSL simultaneously. Ideally, we would like an FSL solution where the server computation and storage do not scale with the number of participating clients.

In this paper, we propose a novel federated split learning framework, termed \alg, that addresses these key challenges by making the communication and storage requirements more efficient, enabling broader adoption of FSL in resource-constrained local devices. 
\alg not only reduces the massive communication cost but also saves storage by keeping a \emph{single} server-side model shared across clients \emph{regardless of the number of clients}. By adding an auxiliary network, the client-side model can be updated locally without waiting for the gradients from the server. Specifically, for a given mini-batch of data, the client does not need to communicate per-batch forward signals with the server in every forward-backward pass to complete the gradient computation. This relaxation has significant practical benefits, as now the clients can operate in an asynchronous fashion that is (almost) independent of other clients and the server\footnote{In this work, the server is responsible for both training the server-side model and aggregating client-side updates. While this design simplifies the architecture and aligns with our focus on communication efficiency, it can be extended to settings where these roles are separated across different servers when privacy is of concern.}. 
Before each aggregation, all clients send their locally trained client-side models and auxiliary networks to the server, possibly also in an asynchronous fashion. The received models are aggregated on the server and redistributed again. These aggregated models are used as the initial model for the next round of \alg. Notably, we propose a single server-side model training framework that performs model updates \emph{only when the smashed data from a client arrives}. The server-side model update thus can be triggered whenever smashed data arrives. In other words, \alg naturally supports \emph{asynchronous client communications} and the resulting \emph{event-triggered server-side model update}.\footnote{
Compared to traditional FSL frameworks that maintain a dedicated server-side model for each client, \alg requires only a single shared server-side model. This design significantly reduces the server's storage requirement, enabling \alg to scale efficiently with a large number of clients. Combined with asynchronous updates, \alg is well-suited for real-world applications involving heterogeneous edge devices and dynamic client participation.}
This feature, similar to how the previously mentioned auxiliary network can allow for asynchronous operations of the clients, would allow the server to operate without waiting for all client updates. Our key contributions are summarized as follows:  
\begin{itemize}
    \item We introduce \alg, a federated split learning framework that integrates an auxiliary network to enable local updates and a \emph{single} shared server-side model to reduce storage costs. By eliminating the need for per-batch forward signal exchanges and removing the dependency on server-side gradient computation, \alg decreases the overall delay and reduces the number of signals transmitted in both uplink and downlink, significantly improving communication efficiency.   
    \item We develop an \emph{event-triggered} server-side update mechanism that enables asynchronous client communication, eliminating the need for synchronized updates and reducing overall system delay. This design makes \alg particularly suitable for real-world applications with numerous heterogeneous devices and clients that dynamically join or leave the network.  
    \item We provide a theoretical convergence analysis of \alg under non-convex loss functions, offering insights into key system constraints and hyperparameter interactions.  
    \item We conduct extensive experiments on CIFAR-10 and F-EMNIST, demonstrating that \alg outperforms existing FSL methods in both IID and non-IID settings while achieving significant communication and storage efficiency.  Our results further show that CNN-based auxiliary networks are more effective than MLPs for image-based IoT applications, as they reduce model complexity while preserving performance. Additionally, \alg achieves significant communication and storage advantages, particularly when clients have large training datasets or larger splitting layers relative to the client model.   
\end{itemize}

The remainder of this paper is organized as follows.  Related works are surveyed in Section~\ref{sec:related}. The system model is described in Section~\ref{sec:model}. The proposed \alg method and its convergence analysis are elaborated in Sections~\ref{sec:CSE-FSL} and \ref{sec:con} respectively. Experimental results are given in Section~\ref{sec:sim}, followed by the conclusions in Section~\ref{sec:conc}.

\section{Related Works}
\label{sec:related}

\mypara{Federated learning.} FL relies on clients to collaboratively train a global ML model coordinated by a parameter server while keeping data in the clients \cite{konevcny2016federated,mcmahan2017communication,li2020federated,kairouz2019advances}. FL has received significant research interest, but its adoption in mobile or IoT networks has been slow mainly because these devices are heterogeneous and highly resource-constrained. Existing FL solutions require high computational resources on the clients when training a large-scale model, while in practice such a model may not be able to fit in the device's memory and requires heavy computation that the device may not support. 
 Recent approaches have sought to address these challenges by improving communication efficiency, such as using compression or sparsification techniques \cite{lin2017deep,li2020acceleration,wang2018atomo,karimireddy2019error,chen2024iot,Zheng2020jsac,chen2021twc,li2022tits,shen2021commag,wei2024twc}, which can significantly reduce communication overhead. Furthermore, hierarchical FL models, which aim to reduce communication costs by organizing clients into clusters and aggregating updates at different levels, have also shown promise in reducing resource usage \cite{zhu2024efficient}.
FedGSL \cite{zhang2024fedsl} proposes a split layer aggregation algorithm, which improves efficiency by transferring partial model parameters to reduce communication cost. 

\mypara{Split learning (SL).} 
Compared with FL, SL \cite{gupta2018distributed,vepakomma2018split,vepakomma2019reducing} enables learning on resource-constrained clients by splitting the neural network model into two parts and keeping only the first part at the client. Several variants of SL networks have been proposed to handle different tasks, such as in the healthcare domain \cite{vepakomma2018split, vepakomma2019reducing, abuadbba2020can, poirot2019split}, addressing privacy \cite{kim2020multiple, pasquini2021unleashing}, and in IoT networks \cite{matsubara2021neural, palanisamy2021spliteasy, gao2020end}. Moreover, \cite{singh2019detailed} conducts a comparative analysis of the communication efficiency between SL and FL methodologies. Pioneering efforts in this field propose innovative online learning algorithms aimed at identifying the optimal cut layer to minimize the training latency \cite{zhang2021learning}. Furthermore, an extended exploration in \cite{wang2021hivemind} delves into a more intricate SL scheme involving multiple cut layers, leveraging a low-complexity algorithm to select the most optimal set of cut layers for improved performance.

\mypara{Federated split learning (FSL).} 
By combining the advantages of FL and SL, SplitFed \cite{thapa2020splitfed} achieves model training in an SL setup with multiple clients training in parallel as in FL. Previous investigations in FSL have explored its applicability across distinct model architectures, emphasizing the adaptability of the approach. Specifically, FedSL \cite{abedi2023fedsl} delves into FSL implementations on Recurrent Neural Networks (RNN), and \cite{park2021federated} extends this exploration by focusing on FSL applications within the context of Vision Transformers. 
DFL \cite{samikwa2024dfl} proposes a resource-aware split model and dynamic clustering to address the joint problem of data and resource heterogeneity. Additionally, privacy considerations within FSL have been a focal point of research efforts, with \cite{zhang2023privacy, lee2024exploring, lee2024optimizing, zheng2024ppsfl} specifically dedicated to enhancing the privacy aspects of FSL. In the pursuit of efficient on-device learning, FedGKT \cite{he2020group} reformulates FL as a group knowledge transfer training model, significantly reducing the training memory footprint.  
Recent efforts have particularly focused on improving communication efficiency in FSL. EPSL reduces activation gradient dimensions by aggregating last-layer activation gradients \cite{lin2024efficient, zhu2024esfl}.  SplitFC \cite{oh2025communication} proposes communication-efficient split learning frameworks incorporating adaptive feature dropout, quantization, and compression strategies to reduce communication overhead. 
Furthermore, \cite{han2021accelerating} proposes a local loss-based FSL method, which mitigates communication costs by concurrently updating client-side and server-side models, enhancing overall training efficiency. All these works still require the server to have significant storage space to keep \emph{multiple} copies of server-side model updates.  
Additionally, asynchronous update mechanisms, which could improve adaptability to dynamic client availability, have not been adequately explored. 
Notably, there remains significant room for enhancing upstream communications, an area yet to be sufficiently addressed in the current literature.

\begin{table*}[h!]
\scriptsize
\centering
\caption{Notations}
 \setcellgapes{2pt}\makegapedcells
\begin{tabular}{|c|l|c|l|c|l|}
\hline
\textbf{Symbol} & \textbf{Description} & \textbf{Symbol} & \textbf{Description} & \textbf{Symbol} & \textbf{Description} \\ \hline
$\alpha$        & Fraction of client-side model    & $1 - \alpha$   & Fraction of server-side model & $\xv_c$ & Client-side model \\ \hline
$\xv_s$ & Server-side model & $\av_c$ & Auxiliary model & $B$ & Batch size \\ \hline
$w$ & Model parameters & $\zv$ & Data sample & $D_i$ & Local dataset at client i \\ \hline
$\tilde{D_i}$ & A mini-batch of data at client i & $g_{\xv_{c,i}}$ & Smashed data (intermediate layer output) at client i & $q$ & Size of the smashed data \\ \hline
$F_c$ & Client-side loss & $F_s$ & Server-side loss & $T$ & Total communication rounds \\ \hline
$n$ & Number of clients & $h$ & Batch frequency of client-server communication & $C$ & Batch frequency of model aggregations \\ \hline
\end{tabular}%
\label{tab:FSL_notations_2col}
\end{table*}

\section{Federated and Split Learning Paradigms}
\label{sec:model}
We initiate our exploration with an overview of the fundamental optimization problem in FL. Following this, we present the conventional federated and split learning workflows, outlining their standard procedures and mechanisms. Finally, we briefly discuss the limitations of contemporary federated and split learning methods, paving the way for the subsequent elaboration of our proposed \alg. \Cref{tab:FSL_notations_2col} provides an overview of the notations used throughout this paper.

\subsection{The Distributed SGD Problem}
\label{subsec:modelFL}

We focus on the standard empirical risk minimization (ERM) problem in ML, 
\tmcr{
which serves as a fundamental framework for training models by minimizing an expected loss over a given dataset. The goal of ERM is to find an optimal model parameter $\xv$ that minimizes the empirical approximation of the expected loss, given by
\begin{equation} \label{eqn:erm}
\min_{\xv \in \mathbb{R}^d} F(\xv) = \min_{\xv \in \mathbb{R}^d} \frac{1}{|D|} \sum_{\zv \in D} l(\xv; \zv),
\end{equation}
where $\xv \in\mathbb{R}^d$ is the ML model variable that one would like to optimize, $l(\xv; \zv)$ is the loss function evaluated at model $\xv$ and data sample $\zv=(\vect{z}_{\text{in}}, z_{\text{out}})$, which describes an input-output relationship of $\vect{z}_{\text{in}}$ and its label $z_{\text{out}}$, and $F: \mathbb{R}^d\rightarrow\mathbb{R}$ is the differentiable loss function averaged over the total dataset $D$.} We assume that there is a latent distribution $\nu$ that controls the generation of the global dataset $D$, i.e., every data sample $z \in D$ is drawn independently and identically distributed (IID) from $\nu$. We denote
\begin{equation*}
    \xv^* \triangleq \argmin_{\xv \in \mathbb{R}^d} F(\xv), \qquad f^* \triangleq F(\xv^*).
\end{equation*}

General distributed and decentralized ML, including FL, aims at solving the ERM problem \eqref{eqn:erm} by using a set of clients that run local computations in parallel, hence achieving a wall-clock speedup compared with the centralized training paradigm. We consider a distributed ML system with one central parameter server and a set of $n$ clients.  Mathematically, problem \eqref{eqn:erm} can be equivalently written as
 \begin{equation} \label{eqn:erm2}
\min_{\xv \in \mathbb{R}^d} F(\xv) = \min_{\xv \in \mathbb{R}^d} \frac{1}{n} \sum_{i=1}^{n}  F_i(\xv),
\end{equation}
where $F_i(\xv)$ is the local loss function at client $i$, defined as the average loss over its local dataset $D_i$:
\begin{equation} \label{eqn:erm3}
F_i(\xv) \triangleq \frac{1}{|D_i|} \sum_{\zv \in D_i}  l(\xv; \zv).
\end{equation}
We make the standard assumption that local datasets are disjoint, i.e., $D = \cup_{i \in [n]} D_i$. This work focuses on the \emph{full clients participation} setting, where all $n$ clients participate in every round of distributed SGD. To ease the exposition and simplify the analysis, we also assume that all clients have the same size of local datasets, i.e., $|D_i|=|D_j|, \forall i, j \in [n]$.

\subsection{Federated and Split Learning}
\label{subsec:modelFSL}

\begin{figure}
    \centering
    \includegraphics[width=0.45\textwidth]{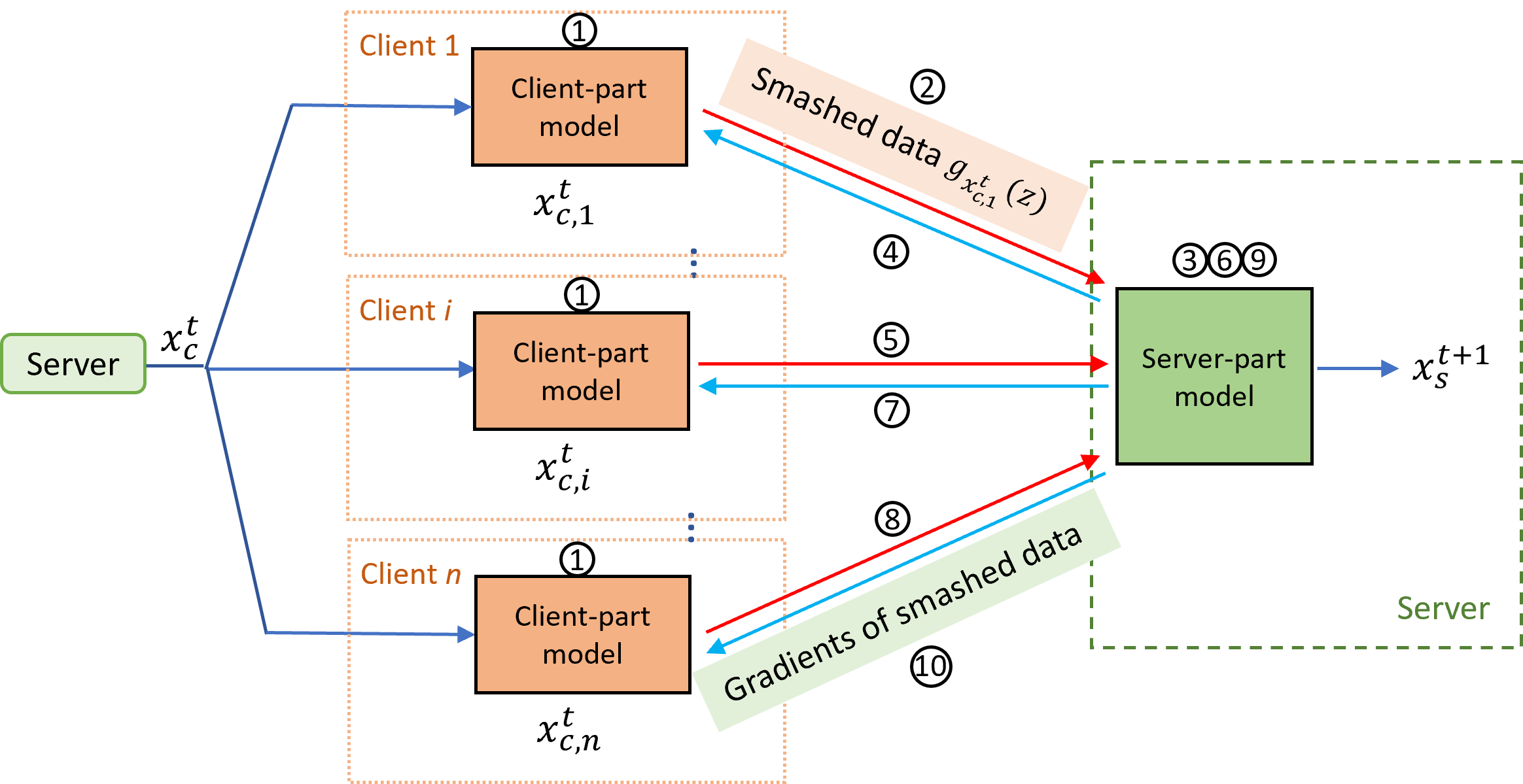}
    \caption{Traditional FSL pipeline in the $t$-th global round.}
    \label{fig:FSLDiagram}
\end{figure}

The FSL pipeline in this paper follows the original framework of \textsc{SplitFed} \cite{thapa2020splitfed}, with an explicit consideration of training client-side models in parallel. The overall workflow for traditional FSL is depicted in \Cref{fig:FSLDiagram}. We first split the complete model $\xv$ into the client-side model $\xv_c$ and the server-side model $\xv_s$. All clients download the client-side model and carry out the forward propagations on their client-side models in parallel (Step \textcircled{1}). They upload their smashed data (Steps \textcircled{2} \textcircled{5} \textcircled{8}) and the corresponding labels to the server. Next, the server continues to process the forward propagation and back-propagation on its server-side model sequentially with respect to the smashed data and then updates the server-side model (Steps \textcircled{3}\textcircled{6}\textcircled{9}). For each model update, the server finally sends the gradients of the smashed data (Steps  \textcircled{4}\textcircled{7}\textcircled{10}) to the respective client for her back-propagation and model update. These steps are repeated until all training data is processed once (i.e., one epoch) and the clients upload the updated client-side models to the server. After the server receives the client-side models from all clients, it aggregates them to generate a new global model and then redistributes the client-side models to clients. FSL then moves to the next round.

\subsection{Challenges}
\label{subsec:largecost}
FSL enables training the client-side models \emph{in parallel}, which particularly improves the efficiency of local model training compared with SL. However, prevalent FSL methods encounter an issue of substantial communication costs. This challenge arises due to the necessity for the server to gather smashed data from all participating clients for each mini-batch data sample (upstream), and the subsequent reliance of all clients on the server's gradients to update their respective models (downstream). 
As a result, the communication overhead for both upstream and downstream requires significant communication resources at each global round. While FL can alleviate this issue by adjusting the communication frequency (e.g., FedAvg \cite{mcmahan2017communication}), \textbf{infrequent communication cannot be naturally implemented in FSL since the client model update needs the backpropagated gradients from the server.} Additionally, FSL still requires the server with extensive storage space to keep and update multiple copies of server-side models, further complicating the implementation in practice. 

\section{Communication and Storage Efficient Federated Split Learning}
\label{sec:CSE-FSL}

To reduce the communication overhead of FSL, a local loss-based training method for SL was proposed in \cite{han2021accelerating}. The key idea is to update client-side and server-side models in parallel by adding an auxiliary network, so the clients can avoid waiting for gradients from the server. However, \cite{han2021accelerating} only improves the downstream communication efficiency, and it still requires the server to have huge storage space to keep multiple copies of the server-side model updates, one for each client, which leads to scalability issues.  Motivated by this issue, we propose a communication and storage efficient federated split learning technique, termed \alg, that not only reduces \emph{both} upstream and downstream communication costs but also saves storage by keeping a \emph{single} server-side model regardless of the number of clients.

In the following, we will first introduce the adopted auxiliary network and the client-side and server-side loss functions, and then present the detailed design of \alg.

\subsection{Auxiliary Network}
\label{subsec:auxiliary}
In the conventional FSL method, the client-side model is updated with the backpropagated signals from the server. In fact, the signals are the gradients of the smashed data (activations), which are produced by calculating the loss from the server-side model. In \cite{han2021accelerating}, the authors propose to add an auxiliary network $\av_c$ to the end of the client-side model and calculate the local loss. In other words, the output of the client-side model is the input of the auxiliary network. 
\tmcr{
The auxiliary network is designed to reconstruct an appropriate output structure for local loss computation. For instance, in classification tasks, the final layer must align with the number of target classes to compute a valid loss value. Since the cut layer is typically positioned in the middle of the network, its output alone is insufficient for direct supervision. The auxiliary network fills this gap by transforming the intermediate features into task-appropriate outputs, thereby enabling local training. This allows clients to update their models independently, without waiting for gradients from the server.
}
For the model architecture, the authors of \cite{han2021accelerating} chose multi-layer perceptrons (MLPs) as auxiliary networks. However, in our study, we have extended this exploration by integrating convolutional neural networks (CNNs) into the framework in a non-trivial way. 
We also compare the size of the auxiliary network under different model architectures. 

For the clients, the goal is to find $\xv_c$ and $\av_c$ to solve ERM problem \eqref{eqn:erm}:
\begin{equation} \label{eqn:erm_c}
\min_{\xv_c, \av_c} F_c(\xv_c) = \min_{\xv_c, \av_c} \frac{1}{n} \sum_{i=1}^{n}  F_{c,i}(\xv_c, \av_c),
\end{equation}
where $F_{c,i}(\xv_c, \av_c)$ is the local loss function at client $i$, defined as the average loss over its local dataset $D_i$:
\begin{equation} \label{eqn:erm_c2}
F_{c,i}(\xv_c, \av_c) =   \frac{1}{|D_i|} \sum_{\zv \in D_i}  l(\xv_c, \av_c; \zv).
\end{equation}

For the server, the proposed scheme in \cite{han2021accelerating} requires the server to have significant storage space because it needs to keep multiple copies of server-side model updates, one for each participating client. Therefore, the goal is to find $\xv_s$ that solves ERM problem \eqref{eqn:erm} based on the optimal client-side model $\xv_c^*$ defined in \Cref{eqn:erm_c}:
\begin{equation} \label{eqn:erm_s}
\min_{\xv_s} F_s(\xv_s) = \min_{\xv_s} \frac{1}{n} \sum_{i=1}^{n}  F_{s,i}(\xv_s, \xv_c^*),
\end{equation}
where $F_{s,i}(\xv_s, \xv_c^*)$ is the local loss function of the current server-side model corresponding to the dataset of clients $i$, defined as the average loss over its local dataset $D_i$:
\begin{equation} \label{eqn:erm_s2}
F_{s,i}(\xv_s, \xv_c^*) =   \frac{1}{|D_i|} \sum_{\zv \in D_i}  l(\xv_s ; g_{x_c^*}(\zv)).
\end{equation}
Note that the smashed data of the optimal client-side model $x_c^*$ with input $\zv \in D_i$ is denoted as  $g_{x_c^*}(\zv)$. With the auxiliary network, the clients can update the models locally without waiting for the communication of the gradients of the smashed data.

\subsection{CSE-FSL}
\label{subsec:CSE-FSL}

As in Section~\ref{subsec:auxiliary}, we also consider an auxiliary network in the client-side model. However, we keep only one server-side model instead of multiple replicas to completely decouple the storage requirement from the number of clients. In other words, the amount of storage required at the server does not change with the number of participating clients. Additionally, to further reduce the communication cost, the clients in our method do not upload smashed data in every mini-batch data sample. We denote the number of batches to upload the smashed data as $h$, which means that the server updates the model in every $h$ batches of data. We denote the initial model as $\xv_c^0, \xv_s^0, \av_c^0$. 

\begin{figure}
    \centering
    \includegraphics[width=0.45\textwidth]{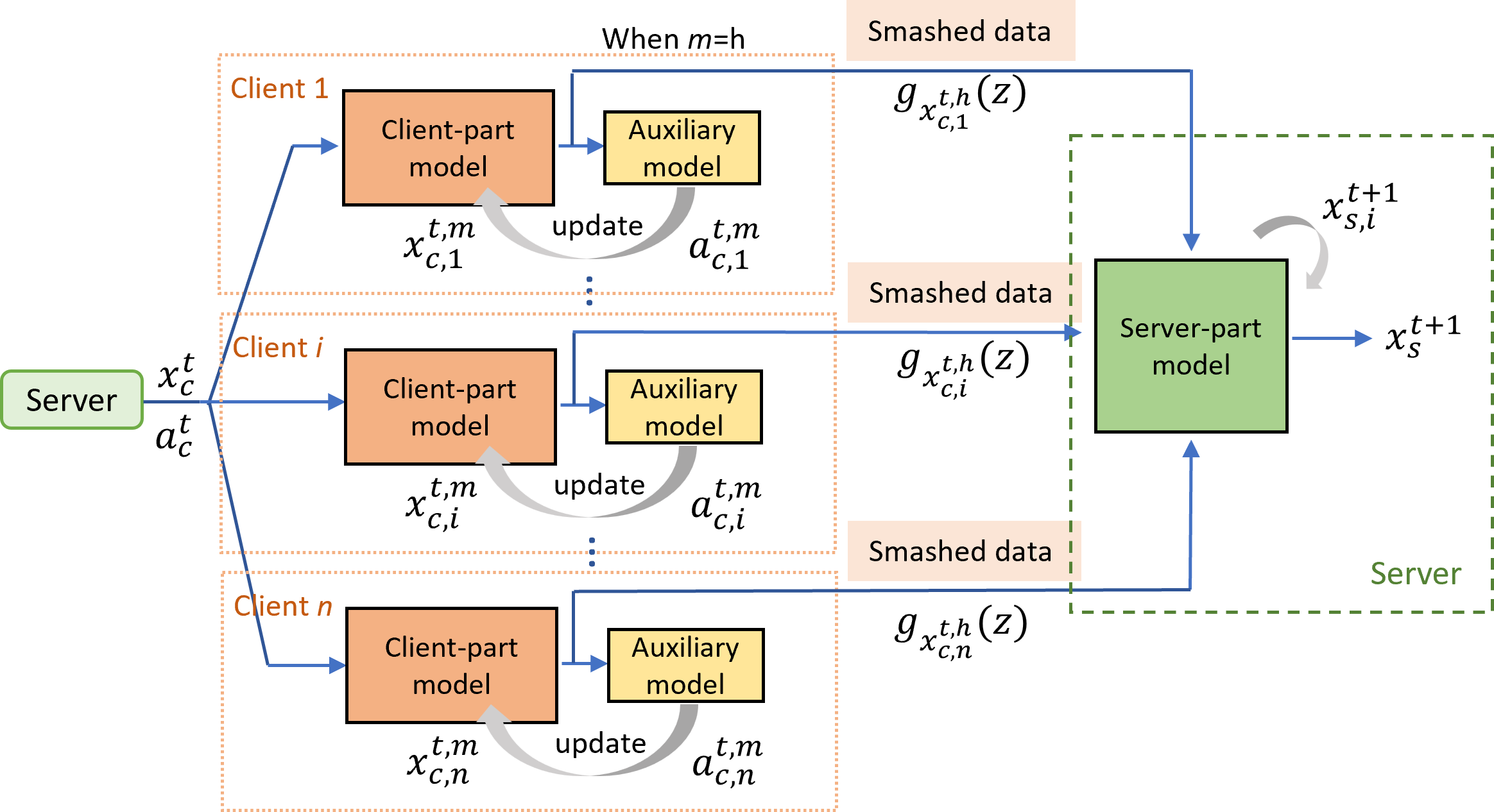}
    \caption{End-to-end \alg pipeline in the $t$-th global round.}
    \label{fig:SystemDiagram}
\end{figure}

The overall system diagram is depicted in \Cref{fig:SystemDiagram}.  In particular, the pipeline works by iteratively executing the following steps at the $t$-th learning round, $\forall t \in [T] \triangleq \sth{1, 2, \cdots, T}$.

\label{sec:algorithm}
\mypara{Step 1: model download.} In the beginning of each global round $t$, each client $i$ downloads the client-side model $\xv_c^t$ and the auxiliary model $\av_c^t$ from the server, and sets $\xv_{c,i}^{t,0} = \xv_c^t, \av_{c,i}^{t,0} = \av_c^t$. 

\mypara{Step 2: feedforward and smashed data upload.} For a mini-batch of training samples $\tilde{D_i} \in D_i$, client $i$ (in parallel but \emph{possibly asynchronously}) performs feedforward to the last layer of the auxiliary network $\av_{c,i}^{t,m}$ based on the client-side model $\xv_{c,i}^{t,m}$. In this process, we can calculate the local loss  $F_{c,i}(\xv_{c,i}^{t,m}, \av_{c,i}^{t,m})$, which will be explained in Step 3, for all training samples $\zv \in \tilde{D_i}$. Note that when $m \bmod h=0$, each client $i$ computes the smashed data $g_{\xv_{c,i}}(\zv)$, which is the output of the client-side model, and uploads the smashed data and the labels corresponding to the batch data to the server asynchronously.

\mypara{Step 3: model update.} Based on the local loss  from Step 2, the client-side model and the auxiliary network can be updated through backpropagation as follows:
\begin{equation} \label{eqn:step3}
\begin{cases}
      \xv_{c,i}^{t,m+1} = \xv_{c,i}^{t,m} - \eta_t \tilde{\nabla}_x F_{c,i}(\xv_{c,i}^{t,m}, \av_{c,i}^{t,m}), \\
      \av_{c,i}^{t,m+1} = \av_{c,i}^{t,m} - \eta_t \tilde{\nabla}_a F_{c,i}(\xv_{c,i}^{t,m}, \av_{c,i}^{t,m}). 
    \end{cases} 
\end{equation}

Let $\xv_{c,i}^{t,0} = \xv_{c,i}^{t}, \xv_{c,i}^{t+1} = \xv_{c,i}^{t,h} $, $\av_{c,i}^{t,0} = \av_{c,i}^{t},  \av_{c,i}^{t+1} = \av_{c,i}^{t,h}$. We can rewrite \Cref{eqn:step3} as 

\begin{equation} \label{eqn:step3_2}
\begin{cases}
      \xv_{c,i}^{t+1} = \xv_{c,i}^{t} - \eta_t \sum_{m=1}^{h} \tilde{\nabla}_x F_{c,i}(\xv_{c,i}^{t,m}, \av_{c,i}^{t,m}), \\
      \av_{c,i}^{t+1} = \av_{c,i}^{t} - \eta_t \sum_{m=1}^{h} \tilde{\nabla}_a F_{c,i}(\xv_{c,i}^{t,m}, \av_{c,i}^{t,m}), 
    \end{cases} 
\end{equation}
where $\eta_t$ is the learning rate at round $t$ and $\tilde{\nabla} F_{c,i}(\xv_{c,i}^{t,m}, \av_{c,i}^{t,m})$ is the derivative of the local loss for a specific mini-batch $\zv \in \tilde{D_i}$: 

\begin{equation} \label{eqn:step3_3}
 \tilde{\nabla} F_{c,i}(\xv_{c,i}^{t,m}, \av_{c,i}^{t,m}) = \frac{1}{|\tilde{D_i}|} \sum_{\zv \in \tilde{D_i}}  \nabla l(\xv_{c,i}^{t,m}, \av_{c,i}^{t,m} ; \zv).
\end{equation}
\begin{figure}
    \centering
    \includegraphics[width=0.45\textwidth]{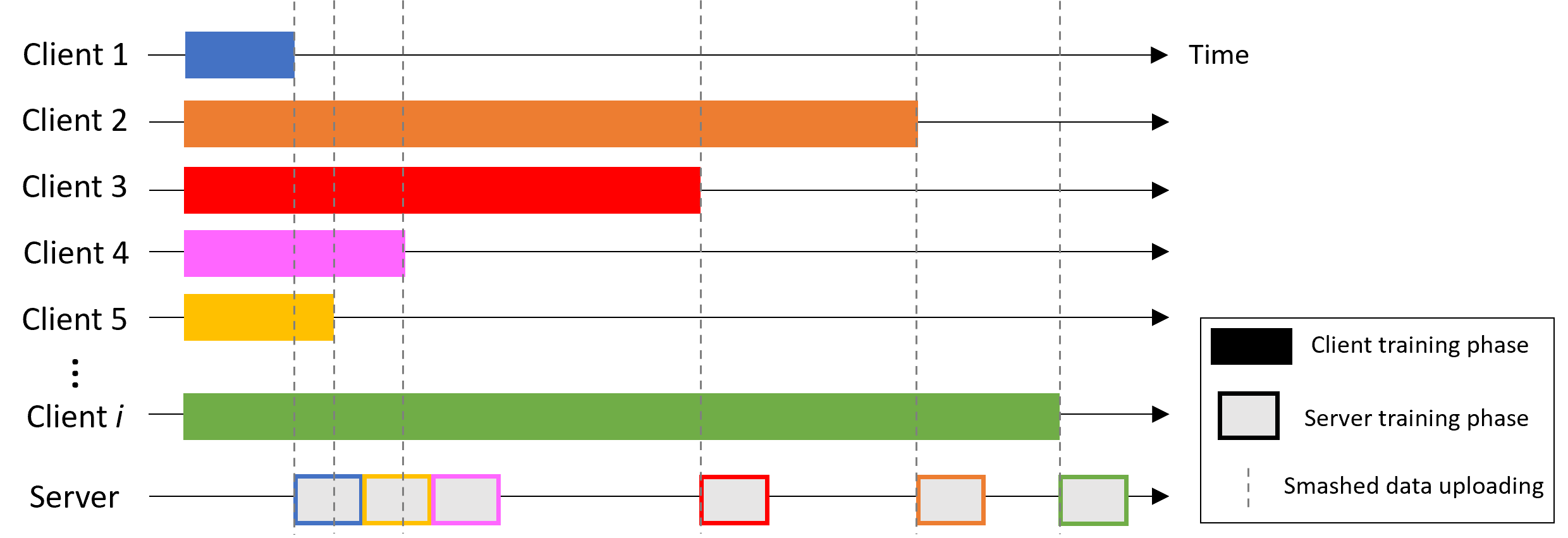}
    \caption{The timeline of asynchronous server training. Each colored box for a client indicates its local training, which can be asynchronous. When the local training is over, the client starts the upload transmission of smashed data. Once such data is received at the server, it immediately begins a server-side model update if it has not started already, treating the just-received data as a mini-batch. }
    \label{fig:async}
\end{figure}

For the server-side model update, the server proceeds feedforward, calculates the loss, and updates the model sequentially when receiving the smashed data $g_{\xv_{c,i}}(\zv)$ from the clients.  
\Cref{fig:async}  illustrates an exemplary scenario of asynchronous client training, upload communication, and server updates in our \alg framework. The variations in training and communication delays across client devices result in notable divergence among the arrivals of individual client smashed data at the server. In our new solution, the server initiates its own training process upon the sequential arrivals of the smashed data uploaded from multiple clients, resulting in an \emph{asynchronous training} approach. This asynchronous server-side model update strategy allows for immediate model iteration upon the arrival of any client smashed data, eliminating the necessity of synchronously awaiting the completion of uploads from \emph{all} clients before commencing the server training phase.  
We acknowledge that when a large number of clients upload smashed data concurrently, the server -- despite using only a single model -- may experience temporary processing delays due to the sequential nature of updates. However, in practice, client uploads are typically staggered due to variations in local training time and network latency, which naturally spreads the server's workload over time. 
The benefit of our approach is significant when the number of clients is large, as the previous synchronous training suffers from the straggler problem while \alg does not. 
Existing asynchronous federated learning methods \cite{wang2022asynchronous, zang2024efficient, xu2023energy} often overlook the performance optimization for resource-limited devices that are incapable of training large models. In the case of asynchronous federated split learning \cite{stephanie2023digital, ao2025semi}, designs frequently rely on multiple server-side models, which result in high storage demands and complex aggregation procedures. In contrast, \alg employs a single server-side model, eliminating additional storage needs and simplifying server-side model aggregation, thereby significantly improving efficiency. 
As a result, our approach improves the efficiency of federated split learning by minimizing the server idle time and maximizing the utilization of available resources.

In particular, the server updates the model after every $h$ batches of the client's data training and we define the model after each update with the corresponding smashed data from client $i$ as $\xv_{s,i+1}^{t+1}$. Therefore, the server performs model updates according to 
\begin{equation} \label{eqn:step3_4}
\xv_{s,i+1}^{t+1} = \xv_{s,i}^{t+1} - \eta_t \tilde{\nabla} F_{s}(\xv_{s,i}^{t+1}, \xv_{c,i}^{t,h}),
\end{equation}
assuming without loss of generality that client data arrives in the natural order of $1, 2, \cdots, n$.

Let $\xv_{s,0}^{t+1} = \xv_{s}^{t}$ and $\xv_{s}^{t+1} = \xv_{s,n}^{t+1}$,  we can rewrite \Cref{eqn:step3_4} as 
\begin{equation} \label{eqn:step3_5}
\xv_{s}^{t+1} = \xv_{s}^{t} - \eta_t \sum_{i=1}^{n} \tilde{\nabla} F_{s}(\xv_{s,i}^{t+1}, \xv_{c,i}^{t,h}),
\end{equation}
where
\begin{equation} \label{eqn:step3_6}
\tilde{\nabla} F_{s}(\xv_{s,i}^{t+1}, \xv_{c,i}^{t,h}) =  \frac{1}{|\tilde{D_i}|} \sum_{\zv \in \tilde{D_i}}  \nabla l(\xv_{s,i}^{t+1} ; g_{\xv_{c,i}^{t,h}}(\zv)).
\end{equation}

\mypara{Step 4: global aggregation.} We denote the time interval between aggregations as $C$, and we first focus on the special case of $C=1$. That means that the global aggregation happens after \emph{every} mini-batch SGD step. Before each aggregation, each client $i$ uploads the updated client-side model and the auxiliary model to the server. Then the server aggregates the client-side model and the auxiliary network as follows:
 \begin{equation} \label{eqn:step4}
\begin{cases}
      \xv_{c}^{t+1} = \frac{1}{n} \sum_{i=1}^{n} \xv_{c,i}^{t+1},   \\
      \av_{c}^{t+1} = \frac{1}{n} \sum_{i=1}^{n} \av_{c,i}^{t+1}. 
    \end{cases} 
\end{equation}
After repeating the overall procedure for $T$ global rounds, the final aggregated model is the integration of the aggregated client-side model and the final server-side model, which is utilized at the inference stage for different tasks.  We present the detailed design of \alg in \Cref{alg:client} and \Cref{alg:server} for client-side and server-side operations, respectively. 

\RestyleAlgo{ruled}
\SetAlFnt{\footnotesize}
\SetAlCapFnt{\footnotesize}
\SetAlCapNameFnt{\footnotesize}
\SetCommentSty{footnotesize}
\LinesNumbered

\begin{algorithm}
\caption{\alg:  clients in the $t$-th global round}\label{alg:client}
Initial the number of batches of local training $h$, and the global aggregation happens after every $c$ batches of local training, (i.e., $C = c$);\

\For {each client $i\in [n]$ in parallel }
{
    Initial current batch number $m = 0$, download $\xv_c^t$ and $\av_c^t$ from the server\;
    Let $\xv_{c,i}^{t,0} = \xv_{c}^t, \av_{c,i}^{t,0} = \av_{c}^t$\;
    
    \For{each mini-batch $\tilde{D_i} \in D_i$ }
    {
         Forward propagation with $\zv \in \tilde{D_i}$ on model $\xv_{c,i}^{t,m}$ and $\av_{c,i}^{t,m}$\, and compute the local loss $F_{c,i}(\xv_{c,i}^{t,m}, \av_{c,i}^{t,m})$\;
         Update the client-side model and the auxiliary network based on \Cref{eqn:step3}\;
         
          \If{$m \bmod h == 0$}{
            Obtain the smashed data $g_{\xv_{c,i}^{t,h}}(\zv)$\;
            Upload $g_{\xv_{c,i}^{t,h}}(\zv)$ and the corresponding labels to the server\;
            Add them to $\text{dataQueue}$ (explained in \Cref{alg:server}) on the server\;
          }
          \If{$m \bmod c == 0$}{
            Upload the updated models $\xv_{c,i}^{t+1}$ and $\av_{c,i}^{t+1}$ to the server for aggregation\;
          }
          $m = m + 1$
    }
}

\end{algorithm}
\begin{algorithm}
\caption{\alg server}\label{alg:server}
Initial client-side model $\xv_c^0$, auxiliary network $\av_c^0$, server-side model $\xv_s^0$, and an empty queue $\text{dataQueue}$;\

Let $\xv_{s,0}^{1} = \xv_{s}^{0}$\;
\For{any $m $}
{
    \If{$m \bmod h == 0$ \tcp{Once receive the smashed data}}
    {
        \While{$\text{dataQueue}$ is not empty}
        {
            Dequeue $g_{\xv_{c,i}^{t,h}}(\zv)$ and the corresponding labels from $\text{dataQueue}$\;
            Forward propagation and calculate the loss $F_{s}(\xv_{s,i}^{t+1}, \xv_{c,i}^{t,h})$\;
            Update the server-side model by \Cref{eqn:step3_4}\;
        }
    }
    \If{$m \bmod c == 0$ \tcp{Once receive the models}} {
        Client-side global model and auxiliary network updates\;
        $\xv_{c}^{t+1} = \frac{1}{n} \sum_{i=1}^{n} \xv_{c,i}^{t+1} $ \tcp{Aggregation} 
        $\av_{c}^{t+1} = \frac{1}{n} \sum_{i=1}^{n} \av_{c,i}^{t+1} $
    }

}

\end{algorithm}

\section{Performance Analysis}
\label{sec:con}

\subsection{Convergence Analysis}
In this section, we analyze the convergence of \alg with non-convex loss functions and IID local datasets. We mainly present the theoretical results with full device participation and the aggregation happens in every communication round (i.e., $|D| = n, C=1$), which sharpens the focus on highlighting the benefit of \alg and presenting a more streamlined theoretical analysis.

\begin{Assumption} \label{ass:smoo}
The client-side and server-side loss functions are $L$-smooth :  $\| \nabla F_c(\xv;\zv)-\nabla F_c(\yv;\zv) \|\leq L \norm{ \xv-\yv }, \| \nabla F_s(\xv;\zv)-\nabla F_s(\yv;\zv) \|\leq L \norm{ \xv-\yv } $ for any $\xv, \yv \in \mathbb{R}^d$ and any $\zv\in \mathcal{D}$. 
\end{Assumption}

\begin{Assumption} \label{ass:bound}
The expected squared norm of stochastic gradients is uniformly bounded: for the client-side loss, we have
$\Expect\| \nabla l(\xv_{c,i}^{t,m}, \av_{c,i}^{t,m};\zv) \|^2 \leq G_1^2 $, $\forall m$, $i \in [n],  t \in [T]$ and  $\zv\in \mathcal{D}$;  similarly,  the server-side loss satisfies: $\Expect\| \nabla l(\xv_{s,i}^t ; g_{\xv_{c,i}^{t,m}}(\zv)) \|^2 \leq G_2^2 $,
 $\forall m$, $i \in [n],  t \in [T]$ and  $\zv\in \mathcal{D}$.
\end{Assumption}


\Cref{ass:smoo} and \Cref{ass:bound} are standard in the literature \cite{boyd2004convex,li2019convergence,wei2022tccn,yang2023jsac}. Furthermore, if we choose diminishing step sizes $\eta_t = \frac{\eta_0}{1+t}$, it will satisfy two conditions: $\sum_{t} \eta_t = \infty$ and $\sum_{t} \eta_t^2 < \infty$. These will be useful in the proofs of our theoretical results.

In each global round $t$, the input distribution of a specific server-side model is determined by $\xv_{c,i}^{t,m} $ and $D_i$. Let $z_{c,i}^t = g_{\xv_{c,i}^{t,h}}(\zv)$ be the output of the $i$-th client-side model at global round $t$, following the probability distribution $P_{c,i}^t(z)$. Noting that $P_{c,i}^t(z)$ is time-varying,  we let $P_{c,i}^*(z)$ be the output distribution of the $i$-th client-side model with $x_c^*$ and $D_i$. We also define the distance between these two distributions as 
\begin{equation}
    \label{eqn:defdcit}
    d_{c,i}^t = \int \norm{P_{c ,i}^t(\zv) - P_{c,i}^*(\zv)} \,d\zv.
\end{equation}
Based on this time-varying distribution, \Cref{ass:dis} is specific to our problem setting, which is based on a similar work \cite{belilovsky2020decoupled} but in a centralized setup.

\begin{Assumption} \label{ass:dis}
For $d_{c,i}^t$ defined in \Cref{eqn:defdcit}, it satisfies $\sum_i d_{c,i}^t < \infty$.
\end{Assumption}

We now formally state our main theoretical results. Specifically, \Cref{prop:conv_c} establishes the client-side model convergence rate, while \Cref{prop:conv_s} concerns the server-side model, both for \alg.

\begin{Proposition}
\label{prop:conv_c}
For \alg, the following inequality holds for the client-side model under \Cref{ass:smoo,ass:bound} when the client-side learning rate  is set as $\eta_t = \frac{1}{Lh\sqrt{T}}$: 
\begin{align} \label{eqn:conv1}
\frac{1}{T} \sum_{t=1}^{T} \Expect \qth{\norm{\nabla F_c(\xv_c^t)}^2} \leq &\frac{4Lh(F_c(\xv_c^{1}) - F_c(\xv_c^{*}))}{(2h-1)\sqrt{T}} \nonumber \\
&+ \frac{2hG_1^2}{(2h-1)\sqrt{T}}.
\end{align}
\end{Proposition}

\begin{Proposition}
\label{prop:conv_s}
The server-side model of \alg satisfies the following inequality under \Cref{ass:smoo,ass:bound,ass:dis} when the server-side learning rate  is set as $\eta_t = \frac{1}{Ln\sqrt{T}}$:
\begin{align} \label{eqn:conv2}
\frac{1}{T} \sum_{t=1}^{T}  
& \Expect \qth{\norm{\nabla F_s(\xv_s^t)}^2} \leq \frac{4Ln (F_s(\xv_s^1) - F_c(\xv_s^*))}{(2n-1)\sqrt{T}} \nonumber \\
&+ \frac{4 G_2^2}{(2n-1)T} \sum_{t=1}^{T} \sum_{i=1}^{n}d_{c,i}^t 
+ \frac{2n G_2^2}{(2n-1)\sqrt{T}}.
\end{align}
\end{Proposition}

The proofs for these two propositions can be found in the Appendix.  As $T$ grows, the right-hand side of \Cref{prop:conv_c} converges to zero at rate $\mathcal{O} \pth{\frac{1}{\sqrt{T}}}$. This result is consistent with the existing convergence rate scaling in federated learning. Our result, however, establishes that the auxiliary network allows for the client training to avoid waiting for gradients from the server while still achieving the same convergence behavior. As for the server-side model convergence, we have from \Cref{prop:conv_s} that: (i) a similar convergence rate scaling as in \Cref{prop:conv_c} can be established; and (ii) the expected gradient norm accumulates around $ \frac{4 G_2^2}{(2n-1)T} \sum_{t=1}^{T} \sum_{i=1}^{n}d_{c,i}^t $ as $T$ goes to infinity. If we make another assumption that $d_{c,i}^t$ decays with $t$, then this drift term will converge to zero, indicating the convergence of the server-side model. Without such an assumption, we have $\inf_{t \leq T} \Expect \qth{\norm{\nabla F_s(\xv_s^t)}^2} <= \mathcal{O} \left(\frac{1}{T}\sum_{t=1}^{T} \sum_{i=1}^{n}d_{c,i}^t \right )$, which indicates a gap away from the stationary point.

\subsection{Communication and Storage Analysis}

\begin{center}
\begin{table}[!t]
\scriptsize
\caption{Total communication cost and storage analysis of the approaches for one global epoch}

 \centering
 \setcellgapes{2pt}\makegapedcells
 \resizebox{\columnwidth}{!}{%
 \begin{tabular}{|l | c | c| c | c|} 
 \hline 
 \textbf{Method} &\makecell{ \textbf{$n$ clients commun.} \\\textbf{for one global epoch}} &\makecell{\textbf{Commun. for} \\\textbf{upload/download}}  & \makecell{\textbf{Total commun.} \\\textbf{(Sum of Column 1 and Column 2)}} & \makecell{\textbf{Server storage}}  \\ [0.5ex] 
 \hline
FSL\_MC \cite{thapa2020splitfed}& $2nq|D|$ & $2n\alpha|w|$ &$2nq|D|+2n\alpha|w|$ & $n|w|$\\
 \hline
FSL\_AN \cite{han2021accelerating}& $nq|D|$ & $2n\alpha(|w|+|a|)$ &$nq|D|+2n\alpha(|w|+|a|)$ & $n(|w|+|a|)$\\
 \hline
CSE\_FSL\_$h$ & $\frac{nq}{h}|D|$ & $2n\alpha(|w|+|a|)$ &$\frac{nq}{h}|D|+2n\alpha(|w|+|a|)$ & $|w|+|a|$\\
 \hline
\end{tabular}%
}
\label{tab:comm_and_storage}
\end{table}
\end{center}

We provide a detailed analysis of total communication and storage costs for one global epoch in \Cref{tab:comm_and_storage}.
In terms of communication, since clients do not need to wait for gradients from the server after sending the smashed data, the communication cost of FSL\_AN is halved compared to FSL\_MC in the first term. Although FSL\_AN introduces additional overhead from transmitting auxiliary network $|a|$, this cost is minimal compared to the savings in smashed data communication, particularly when \( |a| \ll |w| \)\footnote{For example, \( |a|/ |w| \approx 2.16\%  \) in the experiments.}. Furthermore, CSE\_FSL\_$h$ improves upon FSL\_AN by reducing the communication frequency through the factor $h$, where \( h > 1 \) signifies that smashed data is transferred to the server after more than one local training batch. As a result, CSE\_FSL\_$h$ is the most communication-efficient method among all, especially as $h$ increases, making it a better choice for minimizing communication costs in FSL.

For storage, FSL\_AN requires additional space for the auxiliary network \( |a| \), making the total server storage cost of $n(|w| + |a|)$. Since \( |a| \) is generally small compared to \( |w| \), the storage cost for FSL\_AN is slightly higher than FSL\_MC due to the added auxiliary network. On the other hand, the total server storage for CSE\_FSL\_$h$ is \emph{independent of the number of clients}, as it stores only one server model and the auxiliary network. Since the storage cost of CSE\_FSL\_$h$ is constant and does not depend on $n$, it is significantly more efficient as the number of clients increases.

\section{Experimental Results}
\label{sec:sim}

\subsection{Experiment Setup}
\label{sec:settup}

We have carried out the experiments on two popular real-world datasets: CIFAR-10 \cite{krizhevsky2009learning} and F-EMNIST \cite{caldas2018leaf}. For CIFAR-10, we report the results for IID datasets and full clients participation. To demonstrate the effectiveness of our method, we also report the results for non-IID datasets and partial clients participation for the F-EMNIST dataset.

To demonstrate the effectiveness of our proposed method, we compare it with several established baselines that represent standard approaches and recent advances in federated split learning. FSL\_MC and FSL\_OC \cite{thapa2020splitfed} are representative of conventional FSL designs using multiple or single server-side model copies, while FSL\_AN \cite{han2021accelerating} is a more recent method that incorporates auxiliary networks but at high communication and storage costs.
We compare the following methods in the experiments.  
\begin{itemize}
    \item \textbf{FSL\_MC} \cite{thapa2020splitfed}: A standard FSL setup with multiple server-side model copies, where each client communicates with its own dedicated model on the server.
    \item \textbf{FSL\_OC}  \cite{thapa2020splitfed}: Another standard baseline using a single shared server-side model. However, this setup frequently encounters convergence issues. To address this, we incorporate gradient clipping \cite{pascanu2013difficulty} to mitigate gradient explosion. In contrast, CSE\_FSL avoids such instability by using an auxiliary network to define a local objective, allowing the client to train independently of the server's backpropagated gradients.
    \item \textbf{FSL\_AN} \cite{han2021accelerating}: A prior method that adds an auxiliary network on the client side and maintains multiple server-side model copies. Each client uploads smashed data at every batch, incurring high communication and storage overhead.
    \item \textbf{CSE\_FSL\_$h$}: Our proposed method. We attach an auxiliary network to the client-side model and maintain a single shared server-side model. Each client trains on $h$ local batches before uploading smashed data, significantly reducing communication frequency.  In the experiments, the global aggregation is performed once per epoch ($C=1$).
\end{itemize}
All reported results are averaged over five independent runs for statistical reliability.

\mypara{CIFAR-10.} The training sets are evenly distributed over $N = K = 5$ or $10$ clients. The client-side model has two $5 \times 5$ convolution layers  (both with 64 channels), which are both followed by a $2 \times 2$ max-pooling and a local response norm layer. For the auxiliary model, we try two different architectures: one fully connected layer and one $1 \times 1$ convolution layer with different numbers of channels, and a fully connected layer. Besides, the server-side model consists of two fully connected layers (384 and 192 units respectively) with ReLU activation and a final output layer (10 classes) with a softmax function. The training parameters are: batch size $BS = 50$, epoch $E = 1$, learning rate initially sets to $\eta$  = 0.15 and decays every 10 rounds with rate 0.99.

\mypara{F-EMNIST.}  We use the federated version of the EMNIST dataset, F-EMNIST \cite{caldas2018leaf}, in this experiment. There are 3,500 clients with a total of 817,851 samples. It should be noted that F-EMNIST partitions the images of digits or English characters by their authors, thus the dataset is naturally non-IID since the writing style varies from person to person. We use the model recommended by \cite{reddi2020adaptive}, which is a CNN with two convolutional layers, max-pooling, and dropout on the client side, and a 128-unit linear layer with a final output layer (62 classes) with a softmax function. We set $K = 5, BS = 10, E = 1$ and $\eta  = 0.03$ for training.

\subsection{Accuracy Comparison}
\label{sec:results}
We present the top-1 accuracy results as a function of the amount of communication rounds. When client $i$ sends the smashed data to the server, it completes one communication round. 

\begin{figure}[htbp]
\centering
\includegraphics[width=0.48\linewidth]{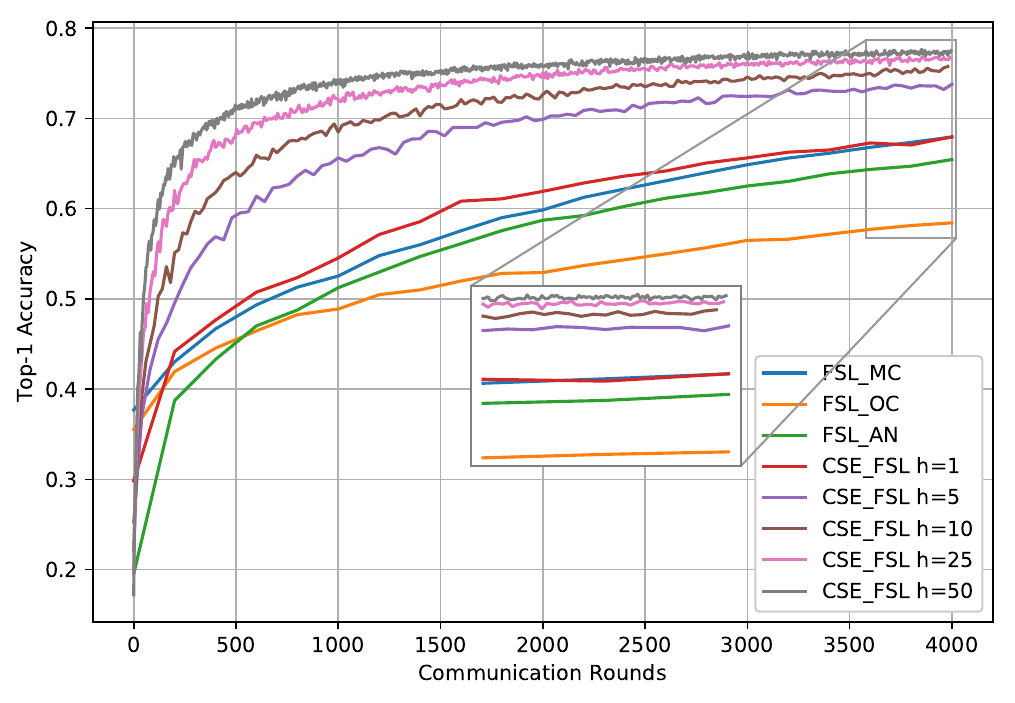}\hfill
\includegraphics[width=0.48\linewidth]{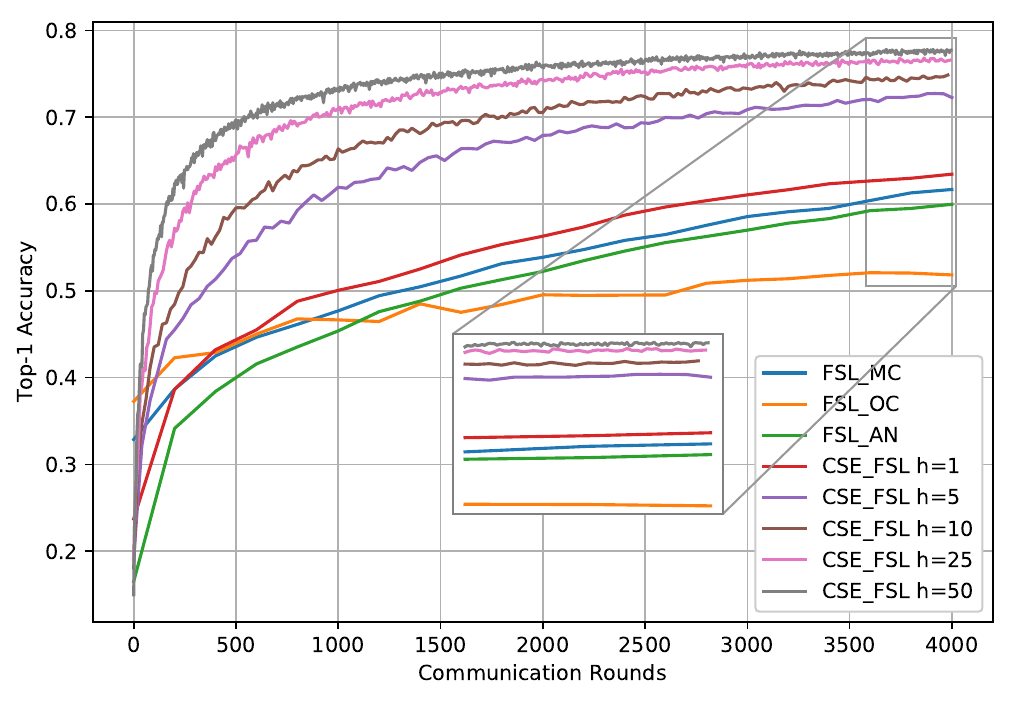}
\hspace*{12pt} \scriptsize (a) 5 clients \hspace{90pt} (b) 10 clients
\caption{CIFAR-10 results with IID local datasets and full clients (5 clients \& 10 clients) participation. }
\label{fig:cifar_5_clients}
\end{figure}

\Cref{fig:cifar_5_clients}(a) shows the performance of each method across communication rounds for IID local datasets with full client (5 clients) participation.
To reduce the downstream communication load, FSL\_AN sacrifices a small amount of performance compared to FSL\_MC. However, with a very small auxiliary network, CSE\_FSL with $h=1$ already outperforms FSL\_OC, despite the latter employing additional ML enhancements while the former does not. This demonstrates that the auxiliary network can address the convergence issue when the server has only a single server-side model. We also evaluate the top-1 test accuracy of our method with varying $h$ values. For instance, $h=5$ means each client locally trains on 5 batches of data before sending the smashed data to the server. Our scheme with a larger $h$ performs better than that with a smaller $h$, as less frequent communication allows the clients to train more between communications and achieve higher accuracy.  
It is evident that \alg outperforms other methods within the same rounds, indicating that our method maintains strong training performance without requiring smashed data uploads for each mini-batch. Overall, the top-1 test accuracy convergence accelerates as $h$ increases, thanks to the additional training steps in each communication round.

A comparative analysis of the results obtained for 10 clients, as shown \Cref{fig:cifar_5_clients}(b), is consistent with the trends observed in \Cref{fig:cifar_5_clients}(a), reinforcing the significant advantages of \alg. As expected, a marginal degradation in performance across epochs was observed with the increase in the number of clients (from 5 to 10). This decline can be attributed to the diminishing amount of data available for individual client training. Consequently, models trained on these reduced and less representative datasets tend to exhibit weaker generalizations, notably impacting performance that is particularly evident in the FSL\_MC, FSL\_OC, and FSL\_AN methods. Remarkably, our proposed \alg demonstrates superior stability in maintaining a nearly identical top-1 accuracy level even in a large number of client scenarios with large $h$. This advantage can be attributed to the client-side model updates by training multiple batches, effectively mitigating the instability. To summarize, the advantages of our proposed \alg are markedly pronounced in scenarios featuring a large number of clients coupled with limited training samples per client.

\begin{figure}[htbp]
\centering
\includegraphics[width=0.48\linewidth]{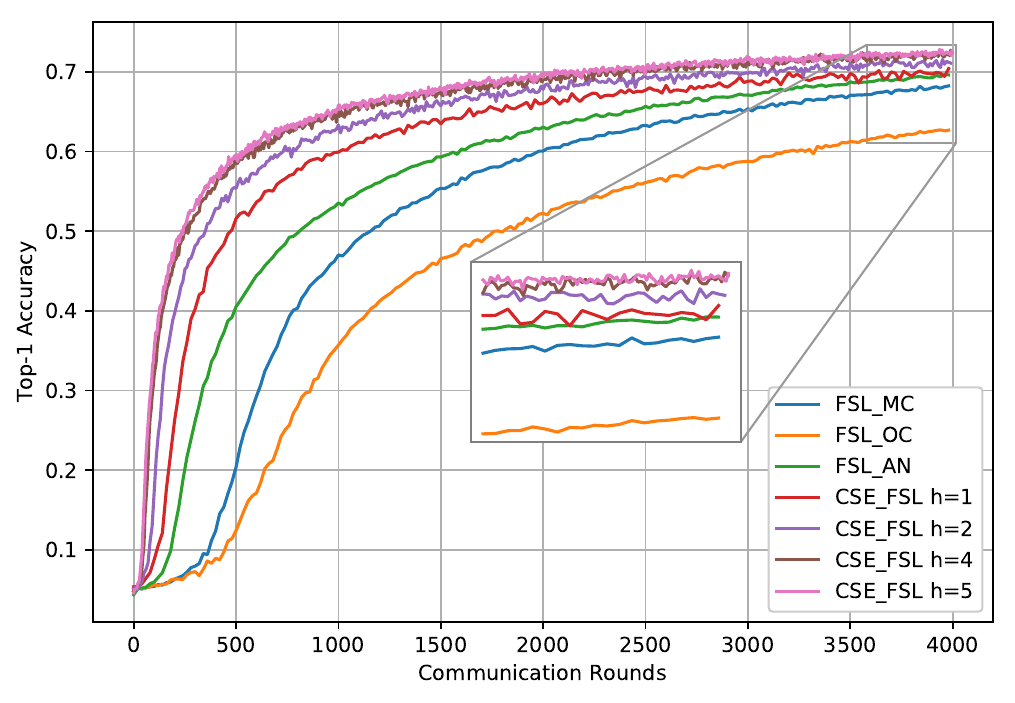}\hfill
\includegraphics[width=0.48\linewidth]{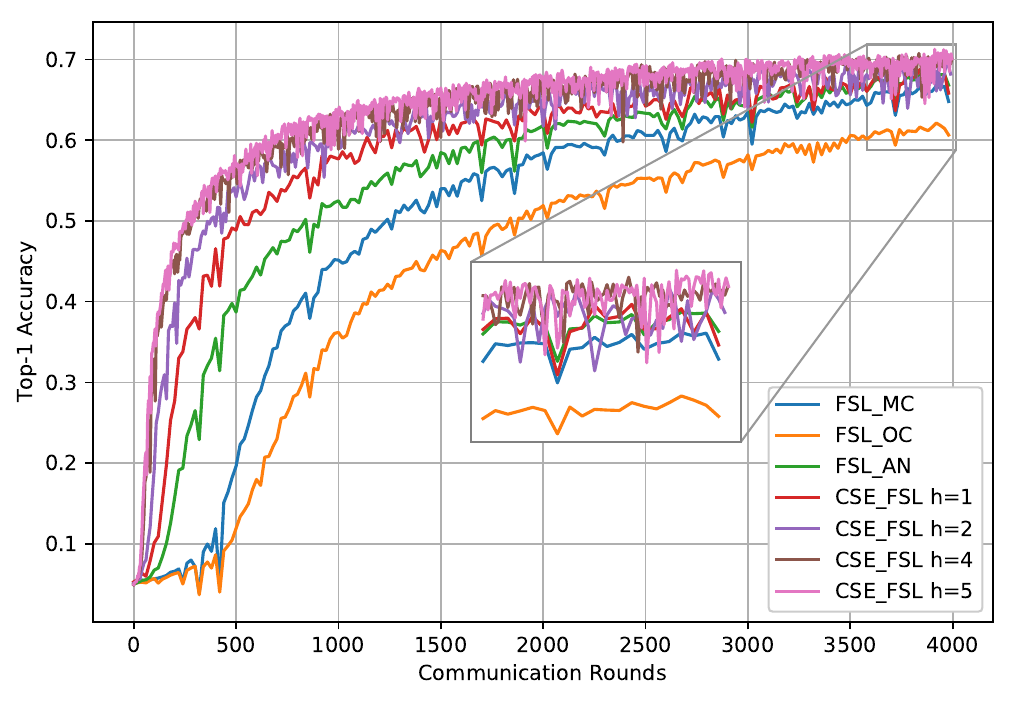}
\hspace*{15pt} \scriptsize (a) IID local datasets \hspace{60pt} (b) non-IID local datasets
\caption{F-EMNIST results with IID and non-IID local datasets and partial clients (5 clients) participation..}
\label{fig:femnist_5_clients_iid}
\end{figure}

In the previous CIFAR-10 experiments, the entire training dataset is partitioned equally among the clients and all the clients participate in the ML training. In the next experiment using the F-EMNIST dataset, we randomly choose partial clients in each global round. We similarly perform model training on the F-EMNIST dataset, and report the results in \Cref{fig:femnist_5_clients_iid}(a) for IID local datasets and partial client participation. We see that FSL\_MC and FSL\_OC perform poorly despite exhaustive parameter tuning, while \alg again converges fast and achieves the top-1 accuracy of 70.42\% to 72.62\% after 4000 communication rounds. Finally, the same conclusion can be drawn from another non-IID experiment using the F-EMNIST dataset, as can be seen in \Cref{fig:femnist_5_clients_iid}(b). After 4000 communication rounds, the accuracy of \alg ranges from 65.85\% to 70.09\%.  
In particular, larger $h$ values lead to faster convergence and higher final accuracy, as extended local training before communication allows clients to better adapt to their non-IID data. This amplifies the performance gap between CSE-FSL and baseline methods under heterogeneous settings. 
This consistent performance across diverse experimental settings further reinforces the resilience and reliability of \alg in addressing challenges related to non-IID data distributions.

\begin{figure}[htbp]
\centering
\includegraphics[width=0.48\linewidth]{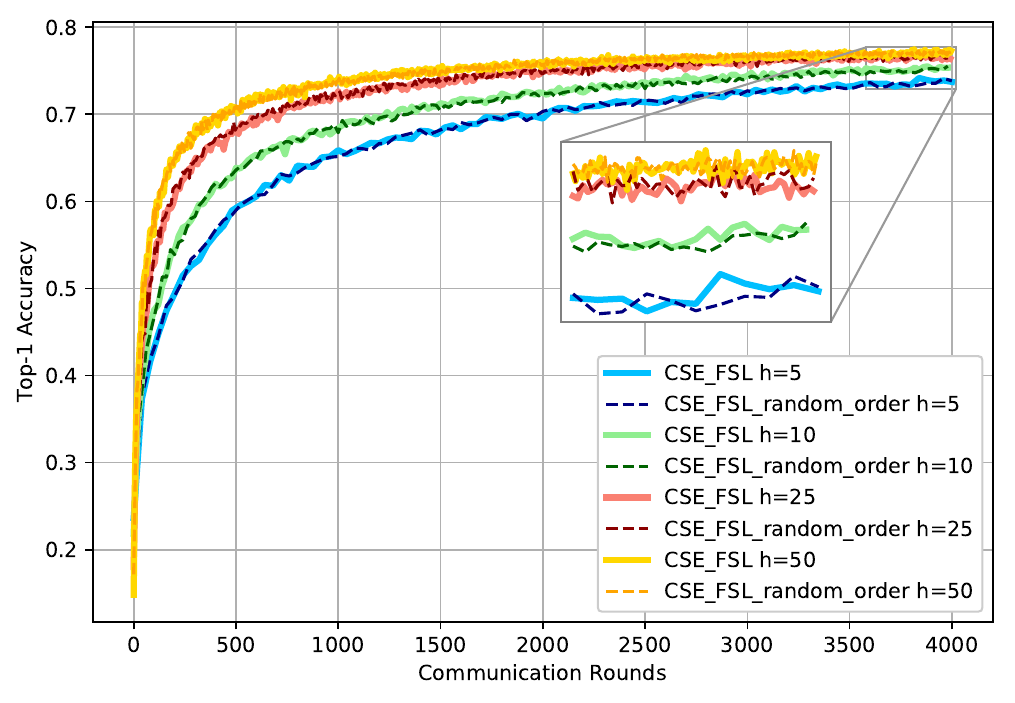}\hfill
\includegraphics[width=0.48\linewidth]{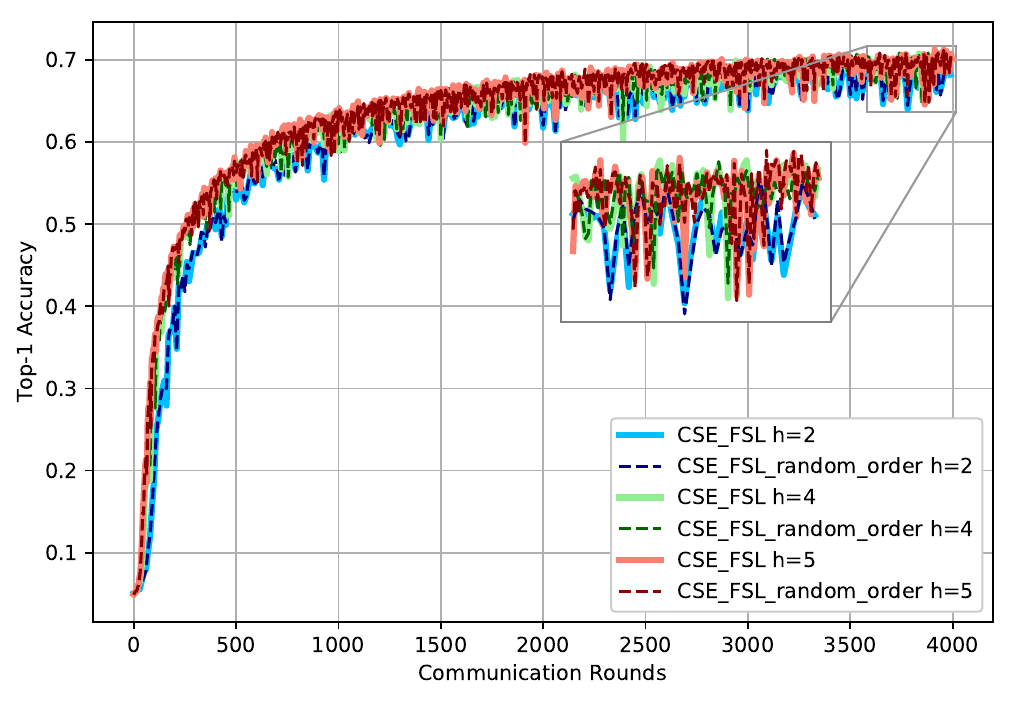}\\
\hspace*{25pt}
\footnotesize (a) CIFAR-10, IID   \hspace{60pt} (b) F-EMNIST, non-IID
\tmcr{
\caption{Comparison of model accuracy under asynchronous server-side training with ordered vs. randomly ordered client updates.}
}
\label{fig:random_order}
\end{figure}

\tmcr{
To validate the effectiveness of our asynchronous server-side model update strategy, we compare the model accuracy under two scenarios: ordered client updates and randomly ordered client updates. As shown in \Cref{fig:random_order}, the resulting accuracies are nearly identical on both CIFAR-10 and F-EMNIST datasets, demonstrating that the update order of client-smashed data does not impact the overall model performance. This confirms that the server can reliably update the global model immediately upon receiving data from any client, regardless of the arrival sequence. These results provide strong empirical support for the practical benefits of our asynchronous design -- specifically, the ability to mitigate the straggler effect and maximize training efficiency without sacrificing accuracy.
}

\subsection{Impact of Auxiliary Network Architectures}
\label{sec:aux}

\begin{center}
\begin{table}[htb]
\scriptsize
\caption{Parameters of auxiliary networks for CIFAR-10 experiments}

 \centering
 \setcellgapes{3pt}\makegapedcells
 \begin{tabular}{|l | c | c | c |} 
 \hline 
 \textbf{Method} &\makecell{ \textbf{Number of}\\ \textbf{output channels}} &\makecell{\textbf{Number of} \\ \textbf{parameters}}  & \makecell{\textbf{Percentage of} \\ \textbf{whole model}}  \\ [0.5ex] 
 \hline
MLP & N/A & 23,050 & 2.16\%\\
 \hline
CNN+MLP & 54 & 22,960 & 2.15\%\\
 \hline
CNN+MLP & 27 & 11,485 & 1.08\%\\
 \hline
CNN+MLP & 14 & 5,960 & 0.56\%\\
 \hline
CNN+MLP & 7 & 2,985 & 0.28\%\\
 \hline
\end{tabular}
\label{tab:parameters}
\end{table}
\end{center}

\begin{figure}[htbp]
\centering
\includegraphics[width=0.48\linewidth]{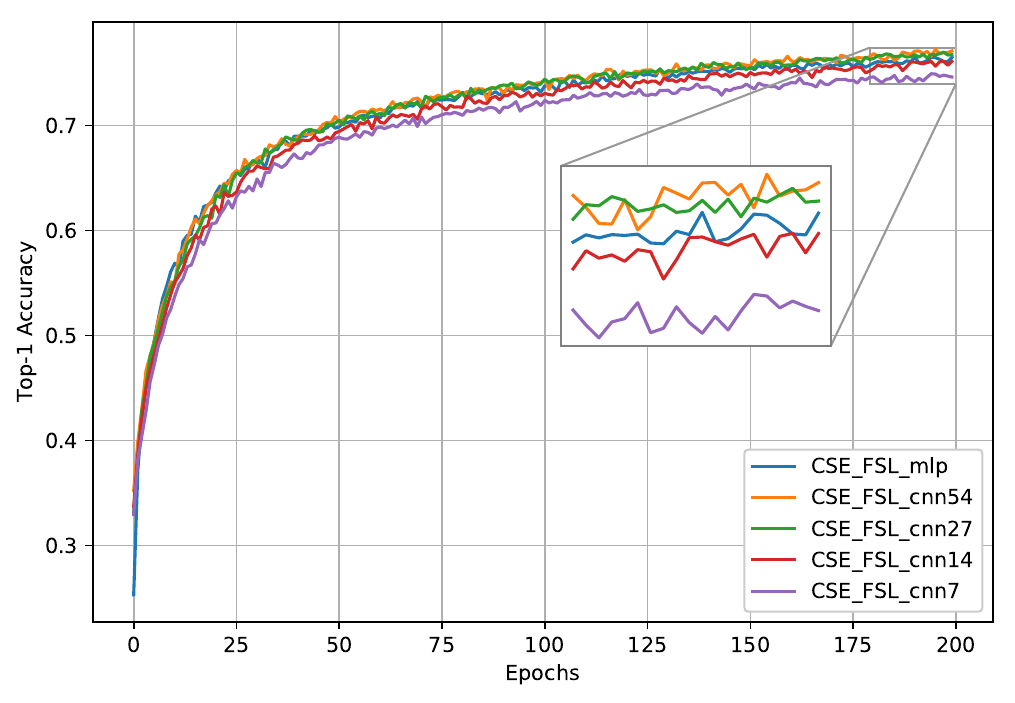}\hfill
\includegraphics[width=0.48\linewidth]{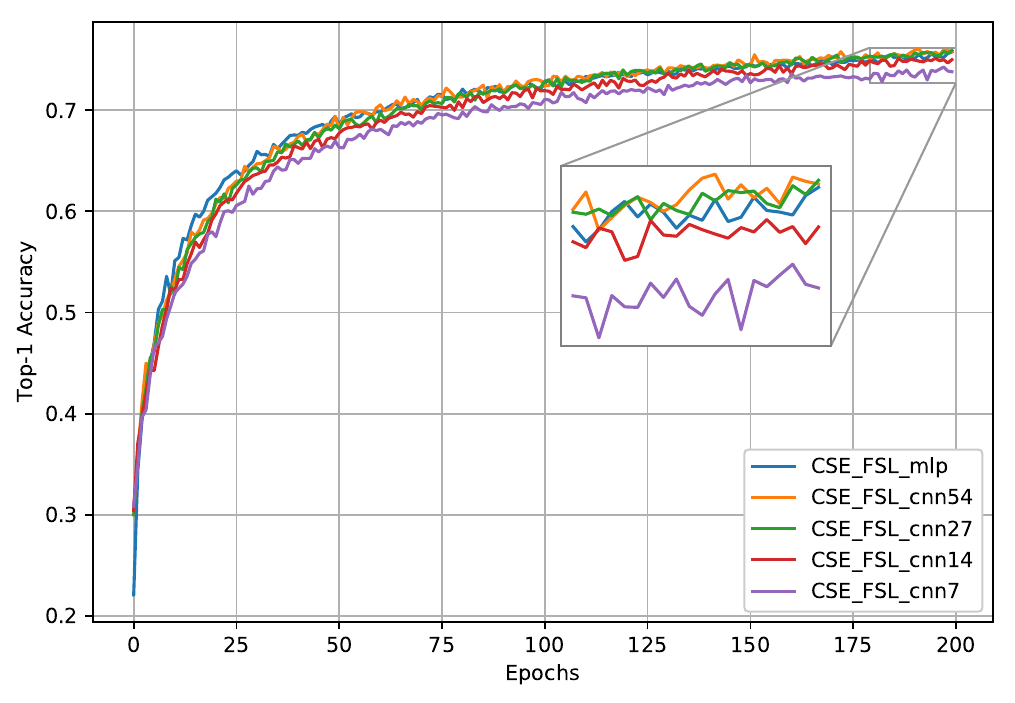}\\
\hspace*{12pt}\footnotesize (a) $h=5$ \hspace{90pt} (b) $h=10$
\caption{CIFAR-10 results with different auxiliary network architectures.}
\label{fig:cifar_5_clients_aux}
\end{figure}

In the CIFAR-10 experiments, the number of model parameters for the client-side model is 107,328, and the server-side model is 960,970. Both CNN and MLP can be utilized as the auxiliary network, and the specific parameters are shown in \Cref{tab:parameters}.

To further explore the impact of different network architectures on the performance, we conduct experiments for IID local datasets and partial clients participation (5 clients). In general, we use $1 \times 1$ convolutional kernels to reduce the dimensionality of the filter space, which avoids the steep dimensionality drop that happens with MLP. Primarily, we study the variation of accuracy with a decrease in the number of output channels of the CNN layer, and \Cref{fig:cifar_5_clients_aux} reports the performance when $h=5$ and $h=10$, respectively. As can be seen from both plots in \Cref{fig:cifar_5_clients_aux}, using a CNN architecture with the same size as an MLP can achieve the same level of top-1 accuracy (77\% in (a) and 76\% in (b)) and convergence rate. However, such performance remains unchanged even when the size of CNN is reduced by a factor of two. The same does not happen to MLP. Accordingly, the auxiliary network with the CNN architecture with 27 channels is much more storage-efficient than the initial MLP since it only needs half of the storage, which is very important for low-power IoT devices.

\begin{center}
\begin{table}[htb]
\scriptsize
\caption{Parameters of auxiliary networks for F-EMNIST experiments}

 \centering
 \setcellgapes{3pt}\makegapedcells
 \begin{tabular}{|l | c | c | c |} 
 \hline 
 Method &\makecell{ Number of\\ output channels} &\makecell{Number of \\ parameters}  & \makecell{Percentage of \\ whole model}  \\ [0.5ex] 
 \hline
MLP & N/A & 571,454 & 47.36\%\\
 \hline
CNN+MLP & 64 & 575,614 & 47.71\%\\
 \hline
CNN+MLP & 32 & 287,838 & 23.86\%\\
 \hline
CNN+MLP & 8 & 72,006 & 5.97\%\\
 \hline
CNN+MLP & 2 & 18,048 & 1.5\%\\
 \hline
\end{tabular}
\label{tab:femnist_parameters}
\end{table}
\end{center}

\begin{figure}[htbp]
\centering
\includegraphics[width=0.48\linewidth]{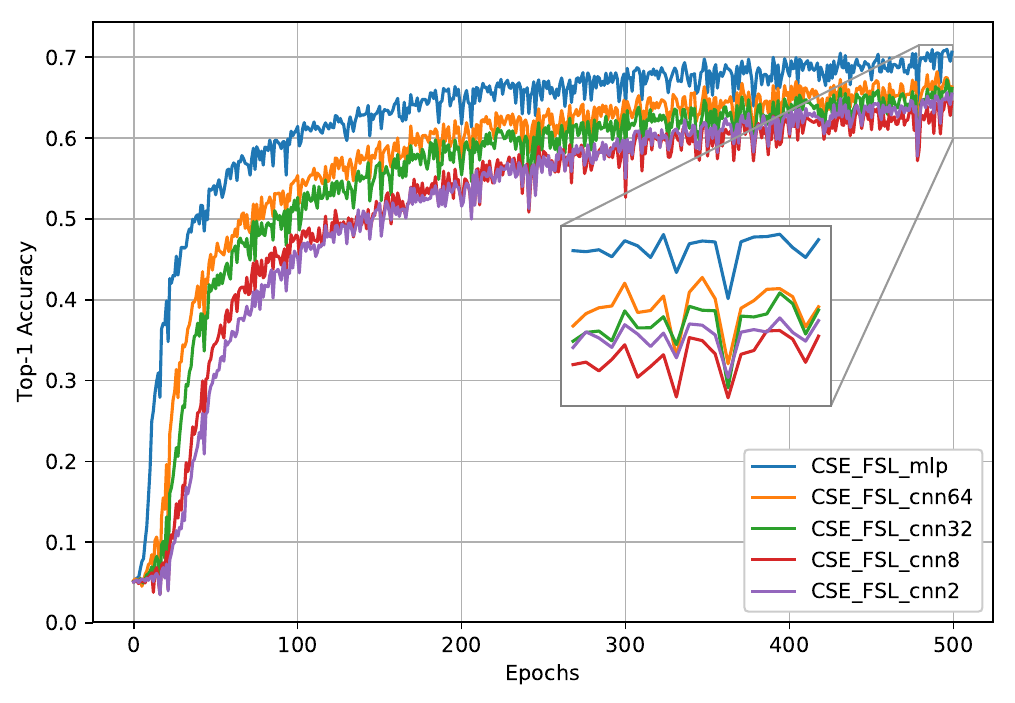}\hfill
\includegraphics[width=0.48\linewidth]{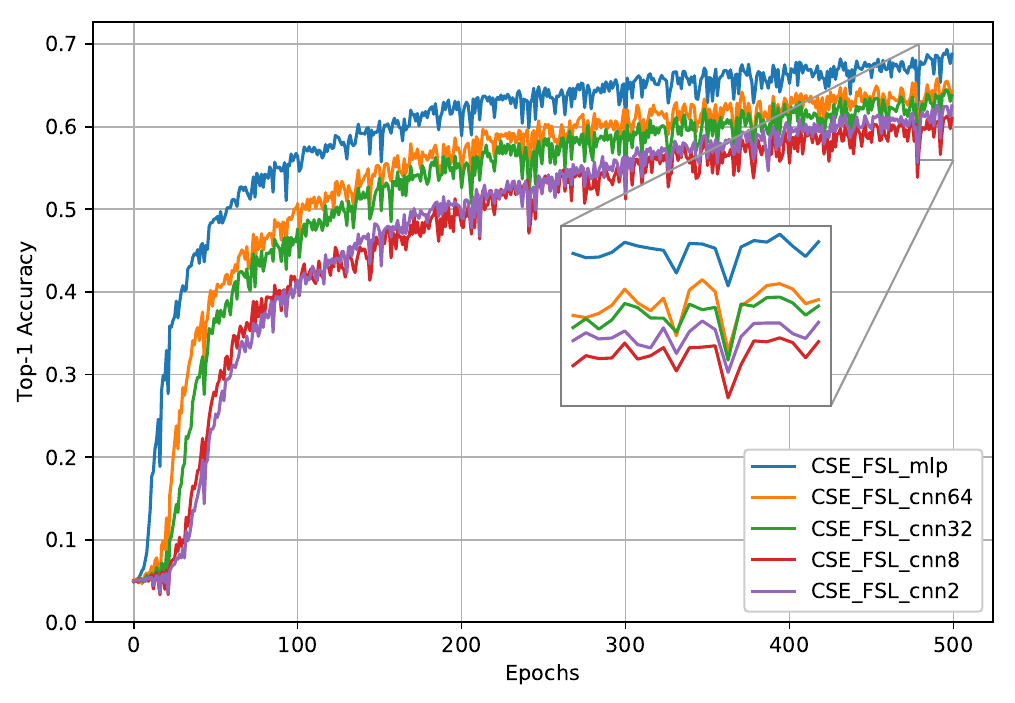}
\hspace*{12pt}\footnotesize (a) $h=2$ \hspace{90pt} (b) $h=4$
\caption{F-EMNIST results with different auxiliary network architectures.}
\label{fig:femnist_5_clients_aux}
\end{figure}

Next, we perform the same experiments for the F-EMNIST dataset, reducing the number of channels from 64 to 2 with parameters shown in \Cref{tab:femnist_parameters}. We only report the results of F-EMNIST with $h=2$ and $h=4$ with non-IID local datasets and partial clients participation in \Cref{fig:femnist_5_clients_aux}, but similar conclusions hold for other conditions. It is known that the classification task of F-EMNIST is harder than CIFAR-10, and one would intuitively expect the ML model to be more complex in this case. With model splitting, the client-side model has 18,816 model parameters, but the server-side model has 1,187,774. We can determine the number of parameters of the auxiliary network with pure MLP architecture to be 571,454, which is significantly larger than that of the client-side model since the number of output neural is 62 (as there are 62 classes). With $1 \times 1$ convolutional kernels, we can reduce the parameters of the auxiliary network by adjusting the output channel, rather than directly cutting down the number of neurons to 62. Although we lose a little of the top-1 accuracy, the purple curve in \Cref{fig:femnist_5_clients_aux} shows that we can achieve training with the same magnitude of client-side model parameters. Therefore, we conclude that when the task is complex (e.g., F-EMNIST), it is better to adopt \alg with a CNN-architectured auxiliary network. Although there is a minor loss of accuracy, the goal of implementing FSL with low communication cost and storage on resource-limited devices can be achieved.

\begin{center}
\begin{table*}[!t]
\caption{Top-1 accuracy, communication load, and storage space comparison at 200 epoch on CIFAR-10 and 500 epoch on F-EMNIST}

 \centering
\adjustbox{max width=1\textwidth}
{
 \setcellgapes{2pt}\makegapedcells
 \begin{tabular}{|l | c | c | c | c | c || c | c | c | c | c |} 
 \hline
\multirow{3}{*}{\diagbox[height=30pt,width=6em]{Methods}{}} & \multicolumn{5}{c||}{CIFAR-10} & \multicolumn{5}{c|}{F-EMNIST} \\
\cline{2-11}

 & & \multicolumn{2}{c|}{Accuracy (\%)} &  \multirow{2}{*}{Load (GB)} & \multirow{2}{*}{Storage (M)}  & &  \multicolumn{2}{c|}{Accuracy (\%)} & \multirow{2}{*}{Load (GB)} & \multirow{2}{*}{Storage (M)} \\ \cline{3-4} \cline{8-9}

 & & IID  & Non-IID & &  & & IID & Non-IID & &  \\
 \hline
FSL\_MC \cite{thapa2020splitfed} &  &\bf{80.55$\pm$0.21} & \bf{61.17$\pm$0.92} & 172.46 & 5.34  &   & 74.52$\pm$0.17 & 72.58$\pm$0.14& 36.23 & 6.03\\
 \hline
FSL\_OC \cite{thapa2020splitfed} & & 73.74$\pm$0.23 &  50.14$\pm$0.44 & 172.46 & \bf{1.50} &  & 70.80$\pm$0.46 & 70.05$\pm$0.16 & 36.23 & \bf{1.28}\\
 \hline
FSL\_AN \cite{han2021accelerating}& & 77.75$\pm$0.10 & 54.67$\pm$1.06 & 86.80 & 5.46 & & \bf{75.05$\pm$0.15} & \bf{73.26$\pm$0.18} & 28.16 & 8.89 \\
 \hline
\multirow{5}{*}{CSE\_FSL} 
& $h=5$ & 76.52$\pm$0.41 & 42.33$\pm$1.02 & \bf{18.14} & \multirow{4}{*}{1.61}  & $h=2$  & 72.01$\pm$0.31 & 70.63$\pm$0.24 & \bf{19.58} & \multirow{4}{*}{4.14}   \\
& $h=10$ & 75.75$\pm$0.53 & 40.24$\pm$1.49 &\bf{9.55} &  &  $h=4$ & 70.27$\pm$0.46 & 68.77$\pm$0.25 & \bf{15.29} &\\
& $h=25$ & 73.57$\pm$0.60  & 37.94$\pm$1.41 & \bf{4.40} & & \multirow{2}{*}{$h=5$}  & \multirow{2}{*}{69.64$\pm$0.50} & \multirow{2}{*}{67.91$\pm$0.58} & \multirow{2}{*}{\bf{14.43}} & \\
& $h=50$ & 73.29$\pm$0.37 & 34.97$\pm$1.04 & \bf{2.69} & & & & &  &\\
 \hline
\end{tabular}
}
\label{tab:com_results}
\end{table*}

\end{center}

\subsection{Communication Load}
\label{sec:comm}

\begin{figure}[htbp]
\centering
\includegraphics[width=0.48\linewidth]{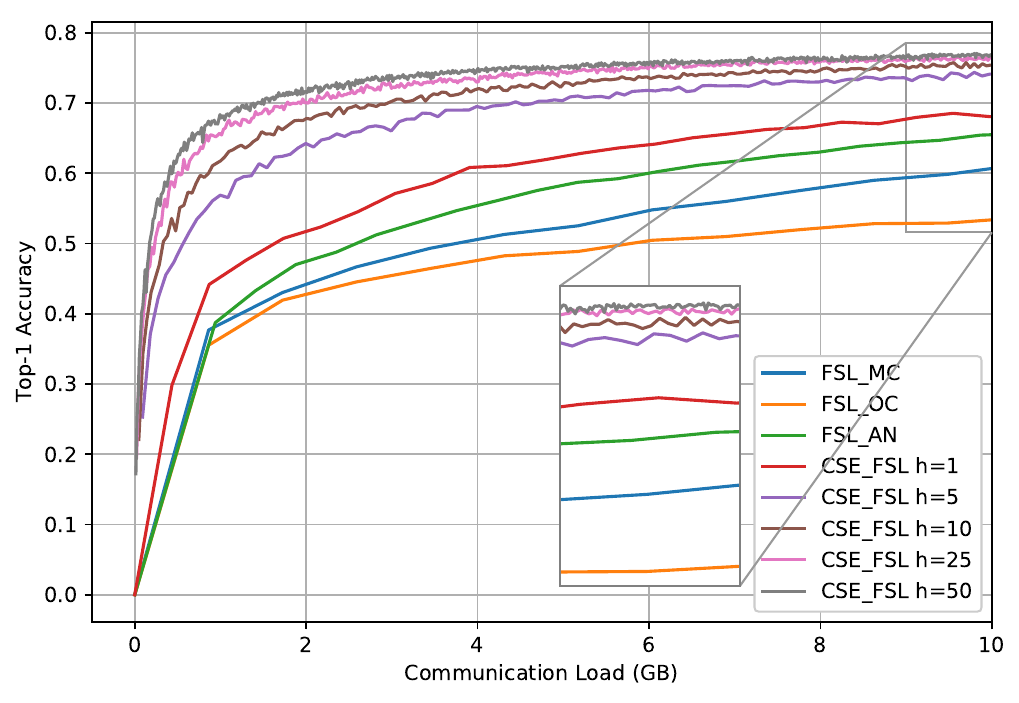}\hfill
\includegraphics[width=0.48\linewidth]{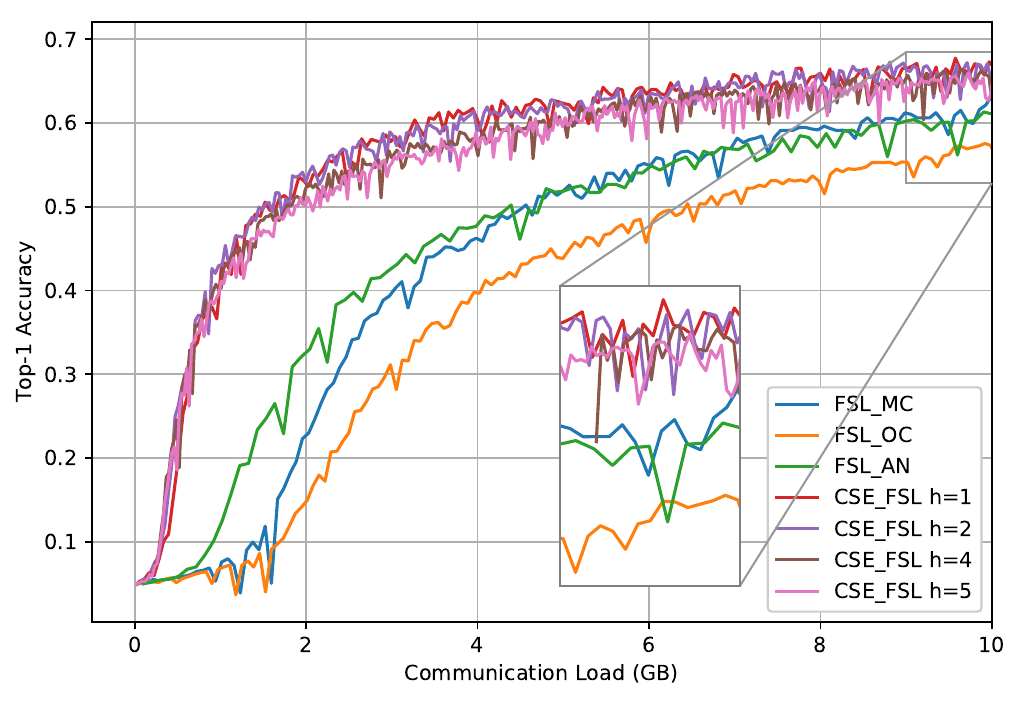}\\
\hspace*{25pt}
\footnotesize (a) CIFAR-10, IID   \hspace{60pt} (b) F-EMNIST, non-IID
\caption{Top-1 test accuracy versus communication load.}
\label{fig:comm}
\end{figure}

Although the previous results have compared the accuracy performance against the number of communication rounds for different methods (including \alg), they ignore the fact that these methods have different loads of communication in each round. To gain more insight into how \alg compares with other methods,  \Cref{fig:comm} depicts the performance of all methods as a function of the communication load under CIFAR-10 IID full clients participation and F-EMNIST non-IID partial clients participation settings. From the left plot, FSL\_AN performs better than FSL\_MC and FSL\_OC since no downlink communication for gradient transfer is required. Compared with FSL\_AN, CSE\_FSL $h=1$ achieves better top-1 accuracy while consuming the same communication load, which shows our proposed CSE\_FSL is much more communication efficient. Furthermore, CSE\_FSL with large $h$ performs better and converges faster than that with small $h$. This is because, in the CIFAR-10 task, the model is relatively simple, but each client has a large number of training samples, and consequently, the total reduction in smashed data uploads accounts for a larger proportion compared with client-side model transfer during the global aggregation.

From the right plot in \Cref{fig:comm}, we can see that all CSE\_FSL methods with different $h$ can reach high top-1 accuracy with less communication load. However, on the F-EMNIST dataset, CSE\_FSL with larger $h$ does not outperform CSE\_FSL $h=1$. This is due to the fact that the auxiliary network is too large (as explained in Section \ref{sec:aux}), and each client is assigned with fewer training samples (due to partial client participation), so the reduction in smashed data is negligible. Therefore, CSE\_FSL with larger $h$ is more suitable for data-heavy clients or the size of the splitting layer is larger relative to the client-side model, because the communication load also depends on the number of mini-batches and the size of each batch transfer.

\subsection{Storage Analysis}
\label{sec:storage}

In the global aggregation steps, the server needs to aggregate the auxiliary networks (if applied) and client-side models. Therefore, the size of storage is proportional to the number of clients. In addition to these, we also need to consider the storage of the server-side model. For example, the method of FSL\_MC has $N$ server-side models, while FSL\_OC and our method CSE\_FSL only need to keep a single server-side model during the whole training process. Here we consider using the total number of parameters to represent the storage size, so the total storage is measured by the sum of the auxiliary network, client-side model, and server-side model sizes. We report the storage in \Cref{tab:com_results}. FSL\_OC uses minimal storage space because it has only one server model and no auxiliary network. FSL\_AN consumes huge storage space due to multiple server-side replicas and auxiliary networks. For our CSE\_FSL, it saves more than 70\% storage space on CIFAR-10 and 53\% storage space on F-EMNIST than FSL\_AN.

\subsection{Comprehensive Analysis}
\label{sec:compre}

In \Cref{tab:com_results}, we also report the top-1 accuracy, communication load, and storage space comparison under different methods on CIFAR and F-EMNIST on both IID and non-IID cases. It summarizes the performance comparisons in Section \ref{sec:results}, Section \ref{sec:comm} and Section \ref{sec:storage}. Compared with FSL\_MC, FSL\_OC reduces storage space, but generally degrades accuracy. Moreover, FSL\_AN improves communication efficiency but incurs higher storage costs. Putting all factors together, these results collectively demonstrate that CSE\_FSL consistently outperforms all other methods when considering the trade-off between top-1 accuracy, communication load, and storage space. In particular, CSE\_FSL clearly outperforms FSL\_AN with higher accuracy, lower communication load, and less storage cost on the CIFAR-10.

\section{Conclusions}
\label{sec:conc}

In this paper, we have proposed a novel federated split learning scheme that is efficient in terms of communication cost and storage space, and present a theoretical analysis that guarantees its convergence. The proposed \alg uses an auxiliary network to locally update client-side models, and update a single server-side model sequentially when the smashed data from any client is received, rather than updating the model for each batch of data or waiting for all clients to finish uploading. Experimental results showed that \alg significantly outperforms existing FSL solutions with a single model or multiple copies in the server. However, challenges remain in deploying \alg in practical settings, such as optimizing model split points to accommodate diverse network conditions and hardware capabilities. 
Additionally, exploring the potential of \alg in more complex real-world scenarios, such as cross-domain applications and intelligent edge computing, presents another promising direction for future research. Investigating how \alg can adapt to heterogeneous data distributions and dynamic environments will be crucial for broader applicability. These issues persist and are left for future exploration.

\bibliographystyle{IEEEtran}
\bibliography{FedLearn,reference,Shen}

\vfill 

\begin{IEEEbiography}[{\includegraphics[width=1in,height=1.25in,clip,keepaspectratio]{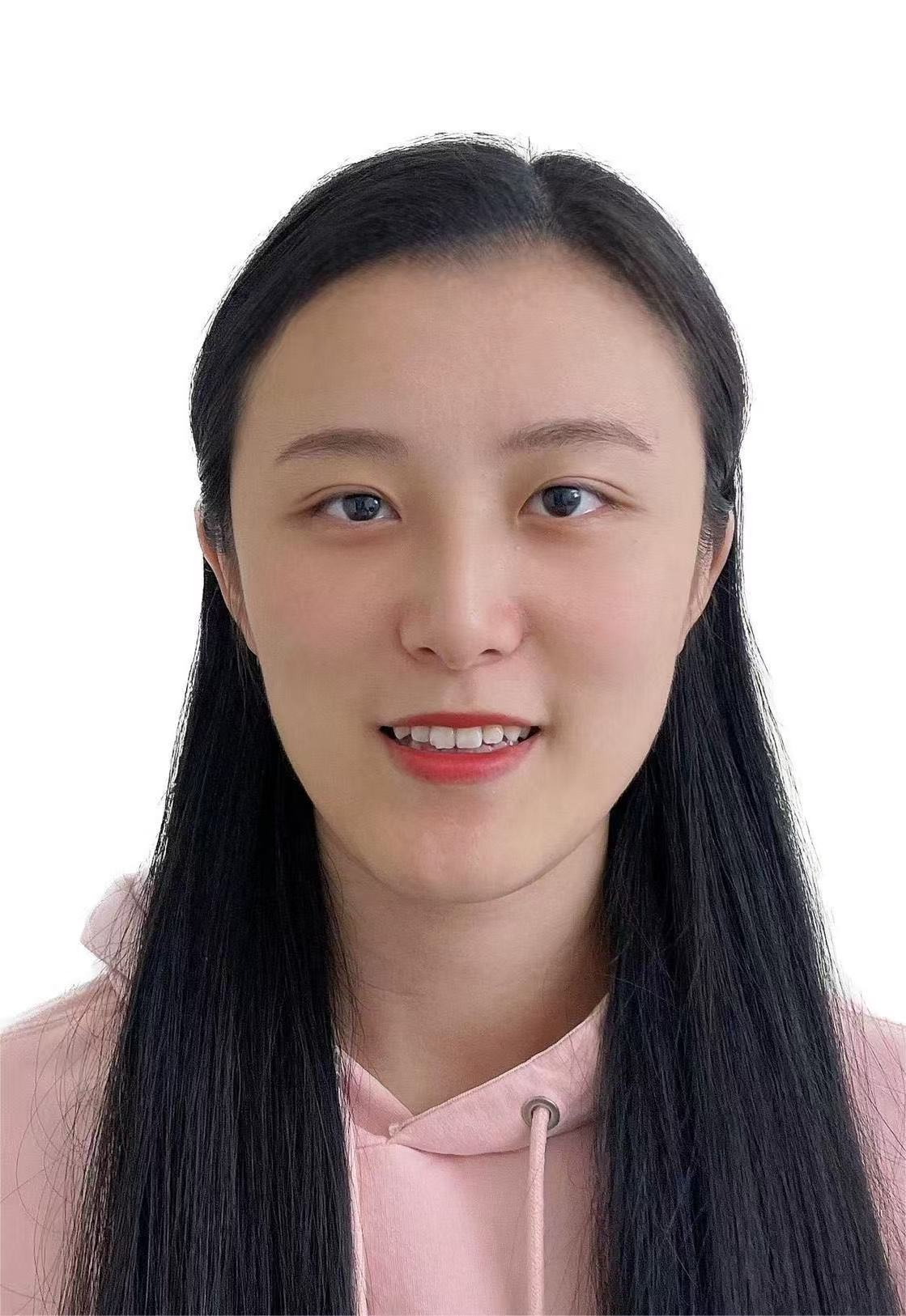}}]{Yujia Mu}
received her B.E. degree from Tsinghua University in 2017 and M.S. degree from the University of Virginia in 2019, where she is currently pursuing a Ph.D. degree in Computer Engineering. Her research interests include federated learning and federated split learning, with a focus on privacy-preserving and distributed machine learning systems.
\end{IEEEbiography}

\begin{IEEEbiography}[{\includegraphics[width=1in,height=1.25in,clip,keepaspectratio]{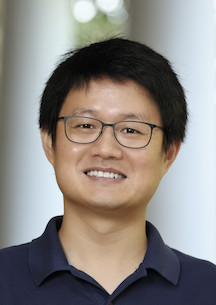}}]{Cong Shen}
(S'01-M'09-SM'15) received his B.S. and M.S. degrees from the Department of Electronic Engineering, Tsinghua University, China. He received the Ph.D. degree in Electrical Engineering from University of California, Los Angeles. He is currently an Associate Professor at the Charles L. Brown Department of Electrical and Computer Engineering, University of Virginia. He also has extensive industry experience, having worked for Qualcomm, SpiderCloud Wireless, Silvus Technologies, and Xsense.ai, in various full time and consulting roles. He received the NSF CAREER award in 2022. He was the recipient of the Best Paper Award in 2021 IEEE International Conference on Communications (ICC), and the Excellent Paper Award in the 9th International Conference on Ubiquitous and Future Networks (ICUFN 2017). His general research interests are in the area of communications, wireless networks, and machine learning. He currently serves as an editor for IEEE Transactions on Wireless Communications IEEE Transactions on Green Communications, and Networking, and IEEE Transactions on Machine Learning in Communications and Networking. He is a founding member of SpectrumX, an NSF Spectrum Innovation Center, and currently serves as the Deputy Director of Collaboration.
\end{IEEEbiography}

\vfill 

\clearpage
\appendices

\section{Proof of \Cref{{prop:conv_c}}}
\label{appdx1}

To simplify the notation, we denote $F_{c,i}(\xv_{c,i}^{t,m}) := F_{c,i}(\xv_{c,i}^{t,m}, \av_{c,i}^{t,m})$. Due to the $L$-smoothness assumption of the client-side loss function, we can write:
\begin{align} \label{eqn:pro_c1}
F_c(\xv_c^{t+1}) \leq & F_c(\xv_c^t) + \nabla F_c(\xv_c^t)^T(\xv_c^{t+1} - \xv_c^t) \nonumber \\
&+ \frac{L}{2} \norm{\xv_c^{t+1} - \xv_c^t}^2.
\end{align}
Since the server aggregates the client-side model based on \Cref{eqn:step4} and client-side model update depending on \Cref{eqn:step3_2}, we have:
\begin{align} \label{eqn:pro_c2}
\xv_{c}^{t+1} &= \frac{1}{n} \sum_{i=1}^{n} \xv_{c,i}^{t+1}  \nonumber \\ 
&=\frac{1}{n} \sum_{i=1}^{n}  \left(\xv_{c,i}^{t} - \eta_t \sum_{m=1}^{h} \tilde{\nabla} F_{c,i}(\xv_{c,i}^{t,m})\right)  \nonumber \\ & 
= \xv_{c}^{t} -  \frac{\eta_t}{n} \sum_{i=1}^{n}  \sum_{m=1}^{h} \tilde{\nabla} F_{c,i}(\xv_{c,i}^{t,m}),
\end{align}
where $ \tilde{\nabla} F_{c,i}(\xv_{c,i}^{t,m}) = \frac{1}{|\tilde{D_i}|} \sum_{\zv \in \tilde{D_i}}  \nabla l(\xv_{c,i}^{t,m}; \zv)$. Now we can rewrite \Cref{eqn:pro_c1} as:
\begin{align} \label{eqn:pro_c3}
F_c(\xv_c^{t+1}) \leq & F_c(\xv_c^t) - \eta_t \nabla F_c(\xv_c^t)^T\left(\frac{1}{n} \sum_{i=1}^{n}  \sum_{m=1}^{h} \tilde{\nabla} F_{c,i}(\xv_{c,i}^{t,m})\right) \nonumber \\ 
& + \frac{L}{2} \eta_t^2 \norm{\frac{1}{n} \sum_{i=1}^{n} \sum_{m=1}^{h} \tilde{\nabla} F_{c,i}(\xv_{c,i}^{t,m})}^2.
\end{align}
Taking  expectation on both sides of \Cref{eqn:pro_c3}, we have:
\begin{align} \label{eqn:pro_c4}
\Expect &\qth{F_c(\xv_c^{t+1})} \leq \Expect \qth{ F_c(\xv_c^t)} \nonumber \\
&- \eta_t \underbrace{ \Expect \qth{\nabla F_c(\xv_c^t)^T\left(\frac{1}{n} \sum_{i=1}^{n}  \sum_{m=1}^{h} \tilde{\nabla} F_{c,i}(\xv_{c,i}^{t,m})\right)}}_{\mathrm{(i)}} \nonumber \\
& + \frac{L}{2} \eta_t^2 \underbrace{\Expect \qth{ \norm{\frac{1}{n} \sum_{i=1}^{n} \sum_{m=1}^{h} \tilde{\nabla} F_{c,i}(\xv_{c,i}^{t,m})}^2}}_{\mathrm{(ii)}}.
\end{align}

We now separately analyze these two items in the right-hand side (RHS) of \Cref{eqn:pro_c4} in the following.

\mypara{Bounding $\mathrm{(i)}$.} First, we introduce an intermediary symbol $\Yv = \frac{1}{n} \sum_{i=1}^{n}  \sum_{m=1}^{h} (\tilde{\nabla} F_{c,i}(\xv_{c,i}^{t,m}) - F_{c,i}(\xv_{c,i}^{t,m})))$ to consider the influence of mini batches. We can then derive a lower bound of $\mathrm{(i)}$ as:
\begin{align} \label{eqn:pro_i1}
&\Expect \qth{\nabla F_c(\xv_c^t)^T\left(\frac{1}{n} \sum_{i=1}^{n}  \sum_{m=1}^{h} \tilde{\nabla} F_{c,i}(\xv_{c,i}^{t,m})\right)} \nonumber \\
&=\Expect \qth{\nabla F_c(\xv_c^t)^T\left(\Yv + \frac{1}{n} \sum_{i=1}^{n}  \sum_{m=1}^{h} \nabla F_{c,i}(\xv_{c,i}^{t,m})\right)} \nonumber\\
& \geq \underbrace{ \Expect \qth{\nabla F_c(\xv_c^t)^T \left(\frac{1}{n} \sum_{i=1}^{n}  \sum_{m=1}^{h} \nabla F_{c,i}(\xv_{c,i}^{t,m}) \right)}}_{\mathrm{(B_1)}} \nonumber \\
&- \underbrace{\norm{\Expect \qth{\nabla F_c(\xv_c^t)^T\Yv}}}_{\mathrm{(B_2)}},
\end{align}
For $\mathrm{(B_1)}$, we can rewrite it as 
\begin{align} \label{eqn:pro_i2}
&\Expect \qth{\nabla F_c(\xv_c^t)^T  \left(\frac{1}{n} \sum_{i=1}^{n}  \sum_{m=1}^{h} \nabla F_{c,i}(\xv_{c,i}^{t,m}) \right) } \nonumber\\
&= \sum_{m=1}^{h} \Expect \qth{\iprod{\nabla F_c(\xv_c^t)} {\frac{1}{n} \sum_{i=1}^{n} \nabla F_{c,i}(\xv_{c,i}^{t,m})}}.
\end{align}
For the expectation term in the RHS of \Cref{eqn:pro_i2}, we have:
\begin{align} \label{eqn:pro_i3}
&\Expect \qth{\iprod{\nabla F_c(\xv_c^t)} {\frac{1}{n} \sum_{i=1}^{n} \nabla F_{c,i}(\xv_{c,i}^{t,m})}} \nonumber\\
&\overset{\mathrm{(a)}}{=} \frac{1}{2}\Expect \qth{\norm{\nabla F_c(\xv_c^t)}^2} 
+ \frac{1}{2}\Expect \qth{\norm{\sum_{i=1}^{n} \frac{1}{n}\nabla F_{c,i}(\xv_{c,i}^{t,m})}^2} \nonumber \\
&- \frac{1}{2}\Expect \qth{\norm{\nabla F_c(\xv_c^t) - \frac{1}{n} \sum_{i=1}^{n} \nabla F_{c,i}(\xv_{c,i}^{t,m}) }^2} \nonumber \\
& \geq \frac{1}{2}\Expect \qth{\norm{\nabla F_c(\xv_c^t)}^2} 
+ \frac{1}{2n^2} \sum_{i=1}^{n} \Expect \qth{\norm{\nabla F_{c,i}(\xv_{c,i}^{t,m})}^2} \nonumber \\
&- \frac{1}{2}\Expect \qth{\norm{\frac{1}{n} \sum_{i=1}^{n} \nabla F_{c,i}(\xv_c^t) - \frac{1}{n} \sum_{i=1}^{n} \nabla F_{c,i}(\xv_{c,i}^{t,m}) }^2} \nonumber \\
&= \frac{1}{2}\Expect \qth{\norm{\nabla F_c(\xv_c^t)}^2} 
+ \frac{1}{2n^2} \sum_{i=1}^{n} \Expect \qth{\norm{\nabla F_{c,i}(\xv_{c,i}^{t,m})}^2} \nonumber \\
&- \sum_{i=1}^{n} \frac{1}{2n^2}\Expect \qth{\norm{\nabla F_{c,i}(\xv_c^t) - \nabla F_{c,i}(\xv_{c,i}^{t,m}) }^2} \nonumber \\
& \overset{\mathrm{(b)}}{\geq} \frac{1}{2}\Expect \qth{\norm{\nabla F_c(\xv_c^t)}^2} 
+ \frac{1}{2n^2} \sum_{i=1}^{n} \Expect \qth{\norm{\nabla F_{c,i}(\xv_{c,i}^{t,m})}^2} \nonumber \\
&- \sum_{i=1}^{n} \frac{L^2}{2n^2}\Expect \qth{\norm{\xv_c^t - \xv_{c,i}^{t,m} }^2},
\end{align}
where (a) holds since $\frac{1}{2}[\norm{U - V}^2] = \frac{1}{2} \norm{U}^2 + \frac{1}{2} \norm{V}^2 - \iprod{U}{V} $, and (b) uses \Cref{ass:smoo}. Using the Cauchy-Schwarz inequality, the last item in the right-hand side of \Cref{eqn:pro_i3} can be bounded as:
\begin{align} \label{eqn:pro_i4}
\Expect \qth{\norm{\xv_c^t - \xv_{c,i}^{t,m} }^2}  
&= \Expect \qth{\norm{\eta_t \sum_{j=1}^{m} \nabla F_{c,i}(\xv_{c,i}^{t,j})}^2} \nonumber \\
&\leq \eta_t^2 m \sum_{j=1}^{m}\Expect \qth{\norm{\nabla F_{c,i}(\xv_{c,i}^{t,j})}^2}.
\end{align}
Thus, we have
\begin{align} \label{eqn:pro_i5}
\mathrm{(B_1)} 
&\geq \sum_{m=1}^{h} \Bigg(\frac{1}{2}\Expect \qth{\norm{\nabla F_c(\xv_c^t)}^2} \nonumber \\
&+ \frac{1}{2n^2} \sum_{i=1}^{n} \Expect \qth{\norm{\nabla F_{c,i}(\xv_{c,i}^{t,m})}^2} \nonumber \\
&- \sum_{i=1}^{n} \frac{L^2\eta_t^2 m}{2n^2} \sum_{j=1}^{m}\Expect \qth{\norm{\nabla F_{c,i} (\xv_{c,i}^{t,j})}^2} \Bigg).
\end{align}
Note that since $\tilde{\nabla} F_{c,i}(\xv_{c,i}^{t,m})$ is an unbiased estimator of $\nabla F_{c,i}(\xv_{c,i}^{t,m})$, we have
\begin{equation} \label{eqn:unbia}
\norm{\Expect \qth{\tilde{\nabla} F_{c,i}(\xv_{c,i}^{t,m}) - \nabla F_{c,i}(\xv_{c,i}^{t,m})}} = 0.
\end{equation}

Next we consider $\mathrm{(B_2)}$. We have: 
\begin{align} \label{eqn:pro_i6}
\norm{\Expect \qth{\nabla F_c(\xv_c^t)^T\Yv}} 
&= \norm{\Expect \qth{\Expect \qth{\nabla F_c(\xv_c^t)^T\Yv | \Omega}}} \nonumber \\
&= \norm{\Expect \qth{\nabla F_c(\xv_c^t)^T \Expect \qth{\Yv | \Omega}}} \nonumber \\
&\overset{\mathrm{(a)}}{\leq} \frac{1}{4} \Expect \qth{ \norm{ \nabla F_c(\xv_c^t)^T}^2} +  \Expect \qth{ \norm{\Expect \qth{\Yv | \Omega}}^2} \nonumber \\
&\overset{\mathrm{(b)}}{=} \frac{1}{4} \Expect \qth{ \norm{ \nabla F_c(\xv_c^t)^T}^2},
\end{align}
where (a) is due to $\Expect[\iprod{U}{V}] \leq \frac{1}{4} \Expect[\norm{U}^2] + \Expect[\norm{V}^2] $, and (b) holds because of \Cref{eqn:unbia}.

Then, applying \Cref{eqn:pro_i5} and \Cref{eqn:pro_i6} to \Cref{eqn:pro_i1} leads to
\begin{align} \label{eqn:pro_i7}
& \mathrm{(i)} =  \Expect \qth{\nabla F_c(\xv_c^t)^T \left(\frac{1}{n} \sum_{i=1}^{n}  \sum_{m=1}^{h} \tilde{\nabla} F_{c,i}(\xv_{c,i}^{t,m})\right)} \nonumber\\
& \geq  \sum_{m=1}^{h} \Bigg(\frac{1}{2}\Expect \qth{\norm{\nabla F_c(\xv_c^t)}^2} 
+ \frac{1}{2n^2} \sum_{i=1}^{n} \Expect \qth{\norm{\nabla F_{c,i}(\xv_{c,i}^{t,m})}^2} \nonumber\\
&- \sum_{i=1}^{n} \frac{L^2\eta_t^2 m}{2n^2} \sum_{j=1}^{m}\Expect \qth{\norm{\nabla F_{c,i}(\xv_{c,i}^{t,j})}^2} \Bigg) \nonumber\\
 &- \frac{1}{4} \Expect \qth{ \norm{ \nabla F_c(\xv_c^t)^T}^2} \nonumber\\
& \geq  \left(\frac{2h-1}{4} \right)\Expect \qth{\norm{\nabla F_c(\xv_c^t)}^2} \nonumber\\
&+ \sum_{m=1}^{h} \Bigg(\frac{1}{2n^2} \sum_{i=1}^{n} \Expect \qth{\norm{\nabla F_{c,i}(\xv_{c,i}^{t,m})}^2} \nonumber\\
&- \sum_{i=1}^{n} \frac{L^2\eta_t^2 m}{2n^2} \sum_{j=1}^{m}\Expect \qth{\norm{\nabla F_{c,i}(\xv_{c,i}^{t,j})}^2} \Bigg).
\end{align}

\mypara{Bounding $\mathrm{(ii)}$.} Using the Cauchy-Schwartz inequality leads to

\begin{align} \label{eqn:pro_ii1}
&\Expect \qth{ \norm{\frac{1}{n} \sum_{i=1}^{n} \sum_{m=1}^{h} \tilde{\nabla} F_{c,i}(\xv_{c,i}^{t,m})}^2} \nonumber\\
&\leq \frac{h}{n}\sum_{i=1}^{n} \sum_{m=1}^{h}\Expect \qth{\norm{\tilde{\nabla} F_{c,i}(\xv_{c,i}^{t,m})}^2} \nonumber\\
&\leq \frac{h}{n}\sum_{i=1}^{n} \sum_{m=1}^{h}\Expect \qth{\norm{\frac{1}{|\tilde{D_i}|} \sum_{\zv \in \tilde{D_i}}  \nabla l(\xv_{c,i}^{t,m}, \av_{c,i}^{t,m} ; \zv)}^2} \nonumber\\
&\leq \frac{h}{n}\sum_{i=1}^{n} \sum_{m=1}^{h} \frac{1}{|\tilde{D_i}|} \sum_{\zv \in \tilde{D_i}} \Expect \qth{\norm{ \nabla l(\xv_{c,i}^{t,m}, \av_{c,i}^{t,m} ; \zv)}^2} \nonumber\\
&\leq h^2 G_1^2,
\end{align}
where the last inequality comes from \Cref{ass:bound}. 

Plugging \Cref{eqn:pro_i7} and \Cref{eqn:pro_ii1} into \Cref{eqn:pro_c4}, we obtain:
\begin{align} \label{eqn:pro_c_sum}
&\Expect \qth{F_c(\xv_c^{t+1})} \leq \Expect \qth{ F_c(\xv_c^t)} - \eta_t \left(\frac{2h-1}{4}\right)\Expect \qth{\norm{\nabla F_c(\xv_c^t)}^2} \nonumber\\
&- \eta_t \sum_{m=1}^{h} \Bigg(\frac{1}{2n^2} \sum_{i=1}^{n} \Expect \qth{\norm{\nabla F_{c,i}(\xv_{c,i}^{t,m})}^2} \nonumber\\
&- \sum_{i=1}^{n} \frac{L^2\eta_t^2 m}{2n^2} \sum_{j=1}^{m}\Expect \qth{\norm{\nabla F_{c,i}(\xv_{c,i}^{t,j})}^2} \Bigg) 
+ \frac{L}{2} \eta_t^2 h^2 G_1^2 \nonumber\\
&\leq \Expect \qth{ F_c(\xv_c^t)} - \eta_t \left(\frac{2h-1}{4}\right)\Expect \qth{\norm{\nabla F_c(\xv_c^t)}^2} + \frac{L}{2} \eta_t^2 h^2 G_1^2.
\end{align}
where the last inequality holds because $$h^2\sum_{m=1}^{h} \Expect \qth{\norm{\nabla F_{c,i}(\xv_{c,i}^{t,m})}^2} \geq  \sum_{m=1}^{h} m\sum_{j=1}^{m}\Expect \qth{\norm{\nabla F_{c,i}(\xv_{c,i}^{t,j})}^2}.$$ 

Then summing up both sides of \Cref{eqn:pro_c_sum} over $t = 1, 2, \cdots, T$. and rearranging the terms yield:
\begin{align} \label{eqn:pro_c_sum2}
&\Expect \qth{F_c(\xv_c^{T+1})} 
\leq \Expect \qth{ F_c(\xv_c^1)} \nonumber\\
&- \left(\frac{2h-1}{4} \right) \sum_{t=1}^{T} \eta_t \Expect \qth{\norm{\nabla F_c(\xv_c^t)}^2} + \frac{h^2 G_1^2 L}{2}  \sum_{t=1}^{T} \eta_t^2.
\end{align}
Finally, using $F_c(\xv_c^{*}) \leq \Expect \qth{F_c(\xv_c^{T+1})} $ and plugging in $\eta_t = \frac{1}{Lh\sqrt{T}}$, we can conclude: 

\begin{align} \label{eqn:pro_c_sum3}
\frac{1}{T} \sum_{t=1}^{T} \Expect \qth{\norm{\nabla F_c(\xv_c^t)}^2} \leq &\frac{4Lh(F_c(\xv_c^{1}) - F_c(\xv_c^{*}))}{(2h-1)\sqrt{T}} \nonumber\\
&+ \frac{2hG_1^2}{(2h-1)\sqrt{T}}.
\end{align}
This completes the proof of \Cref{prop:conv_c}.

\section{Proof of \Cref{{prop:conv_s}}}
\label{appdx2}

We similarly use the following simple notation $f_s(\xv_{s,i}^{t,m}) := F_s(\xv_{s,i}^{t,m}, \xv_{c,i}^{t,m})$ and $F_s(\xv_{s,i}^{t,m}) := F_s(\xv_{s,i}^{t,m}, \xv_c^*)$. Due to the $L$-smoothness assumption of the client-side loss function, we can write:

\begin{equation} \label{eqn:pro_s}
F_s(\xv_s^{t+1}) \leq F_s(\xv_s^t) + \nabla F_c(\xv_s^t)^T(\xv_s^{t+1} - \xv_s^t) + \frac{L}{2} \norm{\xv_s^{t+1} - \xv_s^t}^2.
\end{equation}
Since the server only keeps one copy of the server-side model and the server-side model update depends on \Cref{eqn:step3_5}, we have:

\begin{align} \label{eqn:pro_s2}
\xv_{s}^{t+1} &= \xv_{s}^{t} - \eta_t \sum_{i=1}^{n} \tilde{\nabla} F_{s}(\xv_{s,i}^{t+1}, \xv_{c,i}^{t,h}) \nonumber \\
&= \xv_{s}^{t} - \eta_t \sum_{i=1}^{n} \tilde{\nabla} f_{s}(\xv_{s,i}^{t+1}),
\end{align}
where $ \tilde{\nabla} f_{s}(\xv_{s,i}^{t+1}) = \frac{1}{|\tilde{D_i}|} \sum_{\zv \in \tilde{D_i}}  \nabla l(\xv_{s,i}^{t+1}; g_{\xv_{c,i}^{t,h}}(\zv))$. Now we can rewrite \Cref{eqn:pro_s} and take expectations on both sides:

\begin{align} \label{eqn:pro_s3}
\Expect \qth{F_s(\xv_s^{t+1})} \leq &\Expect \qth{F_s(\xv_s^t)} \nonumber \\
&- \eta_t \underbrace{\Expect \qth{\nabla F_s(\xv_s^t)^T\left(\sum_{i=1}^{n} \tilde{\nabla} f_s(\xv_{s,i}^{t+1})\right)}}_{\mathrm{(i)}} \nonumber \\
&+ \frac{L}{2} \eta_t^2 \underbrace{\Expect \qth{\norm{\sum_{i=1}^{n}  \tilde{\nabla} f_s(\xv_{s,i}^{t+1})}^2}}_{\mathrm{(ii)}}.
\end{align}

\mypara{Bounding $\mathrm{(i)}$.} We similarly introduce an intermediary symbol $\Yv =  \sum_{i=1}^{n}  (\tilde{\nabla} f_s(\xv_{s,i}^{t+1}) - f_s(\xv_{s,i}^{t+1})))$ to consider the influence of mini batches. Therefore, we can obtain a lower bound of $\mathrm{(i)}$ as:
\begin{align} \label{eqn:pro_s4}
\mathrm{(i)}
&=\Expect \qth{\nabla F_s(\xv_s^t)^T \left(\Yv + \sum_{i=1}^{n}   \nabla f_s(\xv_{s,i}^{t+1}) \right)} \nonumber\\
& \geq  \Expect \qth{\nabla F_s(\xv_s^t)^T \left(\sum_{i=1}^{n}  \nabla f_s(\xv_{s,i}^{t+1}) \right)} -\norm{\Expect \qth{\nabla F_s(\xv_s^t)^T\Yv}} \nonumber\\
&= \underbrace{ \Expect \qth{\nabla F_s(\xv_s^t)^T \left(\sum_{i=1}^{n}  \nabla F_s(\xv_{s,i}^{t+1})\right)}}_{\mathrm{(B_3)}} \nonumber \\
&+ \underbrace{\Expect \qth{ \nabla F_s(\xv_s^t)^T \left(\sum_{i=1}^{n}  (\nabla f_s(\xv_{s,i}^{t+1}) - \nabla F_s(\xv_{s,i}^{t+1}))\right)}}_{\mathrm{(B_4)}} \nonumber\\
&- \underbrace{\norm{\Expect \qth{\nabla F_s(\xv_s^t)^T\Yv}}}_{\mathrm{(B_5)}}.
\end{align}

For $\mathrm{(B_3)}$, we can rewrite it as:
\begin{align} \label{eqn:pro_s5}
\mathrm{(B_3)}
&= \sum_{i=1}^{n} \Expect \qth{\nabla F_s(\xv_s^t)^T \left(  \nabla F_s(\xv_{s,i}^{t+1}) \right)}.
\end{align}

For the RHS of \Cref{eqn:pro_s5}, we have:
\begin{align} \label{eqn:pro_s6}
&\Expect \qth{\nabla F_s(\xv_s^t)^T \left(  \nabla F_s(\xv_{s,i}^{t+1}) \right)}  \nonumber\\
&= \frac{1}{2}\Expect \qth{\norm{\nabla F_s(\xv_s^t)}^2} 
+ \frac{1}{2}\Expect \qth{\norm{ \nabla F_s(\xv_{s,i}^{t+1})}^2} \nonumber \\
&- \frac{1}{2}\Expect \qth{\norm{\nabla F_s(\xv_s^t) - \nabla F_s(\xv_{s,i}^{t+1}) }^2} \nonumber \\
& \geq \frac{1}{2}\Expect \qth{\norm{\nabla F_s(\xv_s^t)}^2} 
+ \frac{1}{2}\Expect \qth{\norm{\nabla F_s(\xv_{s,i}^{t+1})}^2}\nonumber \\
&- \frac{L^2}{2}\Expect \qth{\norm{\xv_s^t - \xv_{s,i}^{t+1} }^2}.
\end{align}
The last item in the right-hand side of \Cref{eqn:pro_s6} can be bounded as:
\begin{align} \label{eqn:pro_s7}
\Expect \qth{\norm{\xv_s^t - \xv_{s,i}^{t+1} }^2} 
&= \Expect \qth{\norm{\eta_t \sum_{j=1}^{i} \nabla F_s(\xv_{s,j}^{t+1})}^2} \nonumber \\
& \leq \eta_t^2 i \sum_{j=1}^{i} \Expect \qth{\norm{ \nabla F_s(\xv_{s,j}^{t+1})}^2}.
\end{align}
Thus, we have
\begin{align} \label{eqn:pro_s8}
\mathrm{(B_3)} \geq &\sum_{i=1}^{n} \Bigg(\frac{1}{2}\Expect \qth{\norm{\nabla F_s(\xv_s^t)}^2} 
+ \frac{1}{2}  \Expect \qth{\norm{\nabla F_s(\xv_{s,i}^{t+1})}^2}\nonumber \\
&- \frac{L^2\eta_t^2 i}{2} \sum_{j=1}^{i} \Expect \qth{\norm{ \nabla F_s(\xv_{s,j}^{t+1})}^2}\Bigg).
\end{align}

Now we analyze $\mathrm{(B_4)}$. We have
\begin{align} \label{eqn:pro_s9}
\mathrm{(B_4)}
&\geq -\Expect \qth{\norm{\nabla F_s(\xv_s^t)^T(\sum_{i=1}^{n}  (\nabla f_s(\xv_{s,i}^{t+1}) - \nabla F_s(\xv_{s,i}^{t+1})))}} \nonumber \\ 
& =-\Expect \qth{\norm{\nabla F_s(\xv_s^t)^T} \norm{\sum_{i=1}^{n}  (\nabla f_s(\xv_{s,i}^{t+1}) - \nabla F_s(\xv_{s,i}^{t+1})))}} \nonumber \\ 
&\overset{\mathrm{(a)}}{\geq} -G_2 \Expect \qth{\norm{\sum_{i=1}^{n}  (\nabla f_s(\xv_{s,i}^{t+1}) - \nabla F_s(\xv_{s,i}^{t+1})))}} \nonumber \\ 
& \geq -G_2 \sum_{i=1}^{n}  \Expect \qth{\norm{\nabla f_s(\xv_{s,i}^{t+1}) - \nabla F_s(\xv_{s,i}^{t+1}))}} \nonumber \\ 
& = -G_2 \sum_{i=1}^{n}  \Expect \Bigg[\bigg\Vert  \frac{1}{|D_i|} \sum_{\zv \in D_i} \nabla l(\xv_{s,i}^{t+1} ; g_{\xv_{c,i}^{t,h}}(\zv)) \nonumber \\ 
&-  \frac{1}{|D_i|} \sum_{\zv \in D_i} \nabla l(\xv_{s,i}^{t+1} ; g_{\xv_{c}^*}(\zv)) \bigg\Vert\Bigg]\ \nonumber \\ 
& = -G_2 \sum_{i=1}^{n}  \Expect \Bigg[\bigg\Vert \int \nabla l(\xv_{s,i}^{t+1} ;\Zv) P_{c,i}^t(\zv) \,d\zv  \nonumber \\ 
&-  \int \nabla l(\xv_{s,i}^{t+1} ;\Zv) P_{c,i}^*(\zv) \,d\zv \} \bigg\Vert\Bigg] \nonumber \\ 
&\geq  -G_2 \sum_{i=1}^{n}  \Expect \qth{ \int \norm{\nabla l(\xv_{s,i}^{t+1} ;\Zv)} \norm{P_{c,i}^t(\zv) - P_{c,i}^*(\zv)} \,d\zv} \ \nonumber \\ 
&\overset{\mathrm{(b)}}{\geq}  -G_2^2 \sum_{i=1}^{n}  d_{c,i}^t,
\end{align}
where (a) comes from \Cref{ass:bound}, and (b) holds due to \Cref{ass:dis}. 

Next, we consider the bound of the last item $\mathrm{(B_5)}$.
\begin{align}\label{eqn:pro_s10}
\mathrm{(B_5)} & = \norm{\Expect \qth{\nabla F_s(\xv_s^t)^T\Yv}} \nonumber \\
&= \norm{\Expect \qth{\nabla F_s(\xv_s^t)^T \Expect \qth{\Yv | \Omega}}} \nonumber \\
&\leq \frac{1}{4} \Expect \qth{ \norm{ \nabla F_s(\xv_s^t)^T}^2} +  \Expect \qth{ \norm{\Expect \qth{\Yv | \Omega}}^2} \nonumber \\
&= \frac{1}{4} \Expect \qth{ \norm{ \nabla F_s(\xv_s^t)^T}^2}.
\end{align}
Then, plugging \Cref{eqn:pro_s8},  \Cref{eqn:pro_s9}  and \Cref{eqn:pro_s10} to \Cref{eqn:pro_s4} leads to
\begin{align} \label{eqn:pro_s11}
\mathrm{(i)}
& \geq  \pth{\frac{2n-1}{4}}\Expect \qth{\norm{\nabla F_s(\xv_s^t)}^2}\nonumber \\  
&+ \sum_{i=1}^{n} \Bigg(\frac{1}{2}  \Expect \qth{\norm{\nabla F_s(\xv_{s,i}^{t+1})}^2} \nonumber \\
&- \frac{L^2\eta_t^2 i}{2} \sum_{j=1}^{i} \Expect \qth{\norm{ \nabla F_s(\xv_{s,j}^{t+1})}^2}\Bigg)
 -G_2^2 \sum_{i=1}^{n}  d_{c,i}^t.
\end{align}

\mypara{Bounding $\mathrm{(ii)}$.} Using the Cauchy-Schwartz inequality, we have:
\begin{align} \label{eqn:pro_s12}
&\Expect \qth{ \norm{\sum_{i=1}^{n}  \tilde{\nabla} f_s(\xv_{s,i}^{t+1})}^2} \nonumber \\ 
&\leq n \sum_{i=1}^{n} \Expect \qth{\norm{\tilde{\nabla} f_s(\xv_{s,i}^{t+1})}^2} \nonumber\\
&= n \sum_{i=1}^{n} \Expect \qth{\norm{\frac{1}{|\tilde{D_i}|} \sum_{\zv \in \tilde{D_i}}  \nabla l(\xv_{s,i}^{t+1} ; g_{\xv_{c,i}^{t,h}}(\zv))}^2} \nonumber\\
&\leq n \sum_{i=1}^{n} \Expect \qth{\frac{1}{|\tilde{D_i}|} \sum_{\zv \in \tilde{D_i}}  \norm{\nabla l(\xv_{s,i}^{t+1} ; g_{\xv_{c,i}^{t,h}}(\zv))}^2} \nonumber\\
&\leq n^2 G_2^2.
\end{align}

Therefore, \Cref{eqn:pro_s3} can be further derived as:
\begin{align} \label{eqn:pro_s_sum}
\Expect \qth{F_s(\xv_s^{t+1})} 
&\leq \Expect \qth{F_s(\xv_s^t)} - \left (\frac{2n-1}{4} \right )\eta_t \Expect \qth{\norm{\nabla F_s(\xv_s^t)}^2} \nonumber\\
&- \sum_{i=1}^{n} \Bigg(\frac{\eta_t}{2}  \Expect \qth{\norm{\nabla F_s(\xv_{s,i}^{t+1})}^2} \nonumber \\
&- \frac{L^2\eta_t^3i}{2} \sum_{j=1}^{i} \Expect \qth{\norm{ \nabla F_s(\xv_{s,j}^{t+1})}^2} \Bigg) \nonumber\\
&+\eta_t G_2^2 \sum_{i=1}^{n} d_{c,i}^t+ \frac{L}{2} \eta_t^2 n^2 G_2^2 \nonumber \\
& \leq \Expect \qth{F_s(\xv_s^t)} - \left (\frac{2n-1}{4} \right )\eta_t \Expect \qth{\norm{\nabla F_s(\xv_s^t)}^2}\nonumber\\
&+\eta_t G_2^2 \sum_{i=1}^{n} d_{c,i}^t+ \frac{L}{2}  n^2 G_2^2 \eta_t^2.
\end{align}
where the last inequality holds because $$n^2\sum_{i=1}^{n} \Expect \qth{\norm{\nabla F_s(\xv_{s,i}^{t+1})}^2} \geq  \sum_{i=1}^{n} i\sum_{j=1}^{i}\Expect \qth{\norm{\nabla F_s(\xv_{s,j}^{t+1})}^2}.$$ 

Summing up both sides of \Cref{eqn:pro_s_sum} over $t = 1, 2, \cdots, T$, and rearranging the terms yield:
\begin{align} \label{eqn:pro_s_sum2}
&\Expect \qth{F_s(\xv_s^{T+1})} \leq \Expect \qth{ F_s(\xv_s^1)} \nonumber\\
&- \left (\frac{2n-1}{4} \right ) \sum_{t=1}^{T} \eta_t \Expect \qth{\norm{\nabla F_s(\xv_s^t)}^2} \nonumber\\
&+ G_2^2 \sum_{t=1}^{T} \left (\eta_t \sum_{i=1}^{n} d_{c,i}^t +  \eta_t^2 \frac{Ln^2}{2} \right ).
\end{align}

Finally, using ${F_s(\xv_s^{*})} \leq \Expect \qth{F_s(\xv_s^{T+1})} $ and plugging in $\eta_t = \frac{1}{Ln\sqrt{T}}$, we can conclude:
\begin{align} \label{eqn:pro_s_sum3}
\frac{1}{T} \sum_{t=1}^{T}  \Expect & \qth{\norm{\nabla F_s(\xv_s^t)}^2} \leq \frac{4Ln (F_s(\xv_s^1) - F_c(\xv_s^*))}{(2n-1)\sqrt{T}}\nonumber\\
&+ \frac{4 G_2^2}{(2n-1)T} \sum_{t=1}^{T} \sum_{i=1}^{n}d_{c,i}^t + \frac{2n G_2^2}{(2n-1)\sqrt{T}}.
\end{align}
This completes the proof of \Cref{prop:conv_s}.


\end{document}